\pdfoutput=1
\documentclass[11pt]{article}

\newcommand\ntfootnote[1]{%
  \begin{NoHyper}
  \renewcommand\thefootnote{}\footnotetext{#1}%
  \addtocounter{footnote}{0}%
  \end{NoHyper}
}

\usepackage[preprint]{acl}

\usepackage{times}
\usepackage{latexsym}

\usepackage[T1]{fontenc}
\usepackage[utf8]{inputenc}

\usepackage{microtype}

\usepackage{inconsolata}

\usepackage{graphicx}

\usepackage{amsthm}
\usepackage{amsmath}
\usepackage{amssymb}
\usepackage[linesnumbered,ruled,vlined]{algorithm2e}
\usepackage{booktabs}
\usepackage{multirow}
\usepackage{tcolorbox}
\usepackage{enumitem}

\title{MIG: Automatic Data Selection for Instruction Tuning by Maximizing Information Gain in Semantic Space}

\author{\textbf{Yicheng Chen$^{1,2}$, Yining Li$^{1\dag}$, Kai Hu$^{1,3}$, Zerun Ma$^{1}$, Haochen Ye$^{1}$, Kai Chen$^{1\dag}$} \\
{$^{1}$Shanghai AI Laboratory}
{$^{2}$Fudan University} 
{$^{3}$Carnegie Mellon University} \\
\\
\textbf{Project page}: \href{https://yichengchen24.github.io/projects/mig}{https://yichengchen24.github.io/projects/mig}
}

\begin{document}
\maketitle
% \twocolumn[{%
%   \renewcommand\twocolumn[1][]{#1}%
%   \maketitle
%     \vspace{-48pt}
%     \captionsetup{type=figure}
%     \centering
%     \includegraphics[width=0.7\linewidth]{latex/figs/teaser-v2.pdf}
%     \vspace{-8pt}
%     \caption{\small Comparison with different data selection methods on (a) knowledge-based benchmarks and (b) human-preference benchmarks based on Llama3.1-8B~\cite{touvron2023llama} and Tulu3~\cite{lambert2024tulu3}. See details in Sec.~\ref{sec:main-results}.}
%     \label{fig:teaser}
%     \vspace{12pt}
% }]

\ntfootnote{$^{\dag}$ Corresponding Author.}

\begin{abstract}
Data quality and diversity are key to the construction of effective instruction-tuning datasets.
With the increasing availability of open-source instruction-tuning datasets, it is advantageous to automatically select high-quality and diverse subsets from a vast amount of data.
Existing methods typically prioritize instance quality and use heuristic rules to maintain diversity. 
However, this absence of a comprehensive view of the entire collection often leads to suboptimal results.
Moreover, heuristic rules generally focus on distance or clustering within the embedding space, which fails to accurately capture the intent of complex instructions in the semantic space.
To bridge this gap, we propose a unified method for quantifying the information content of datasets. This method models the semantic space by constructing a label graph and quantifies diversity based on the distribution of information within the graph. 
Based on such a measurement, we further introduce an efficient sampling method that selects data samples iteratively to \textbf{M}aximize the \textbf{I}nformation \textbf{G}ain (MIG) in semantic space.
Experiments on various datasets and base models demonstrate that MIG consistently outperforms state-of-the-art methods.
Notably, the model fine-tuned with 5\% Tulu3 data sampled by MIG achieves comparable performance to the official SFT model trained on the full dataset, with improvements of +5.73\% on AlpacaEval and +6.89\% on Wildbench.
%
% Code will be available.
\end{abstract}

% MIG: Automatic Data Selection for Instruction Tuning by Maximizing Information Gain in Semantic Spac

% 0
% (Optional) Instruction tuning is a standard approach to ...

% 1
% Data quality and diversity are considered crucial to building instruction tuning dataset
% Given the emergence of available instruction tuning datasets, it is advantageous to select optimal subsets from a vast amount of data automatically.

% 2
% 已有方法通常将样本质量和分布多样性分开考虑，这种割裂导致 suboptimal 的采样结果
% 并且，通常采用聚类等简单方式提高数据多样性，无法准确表征数据在语义（semantic）和目的（intention）层面的分布

% 3
% 我们提出一种统一考虑质量和分布的信息量度量方法
% 通过 labelgraph 建模语义空间，优化样本在语义空间中的多样性和均衡性

% 4
% 实验结果
\section{Introduction}

% 1. 背景介绍，引入 Data selection in SFT
% 

%--------------------------------- Teaser Image ---------------------------------%
\begin{figure}[t]
    \centering
    \includegraphics[width=0.9\linewidth]{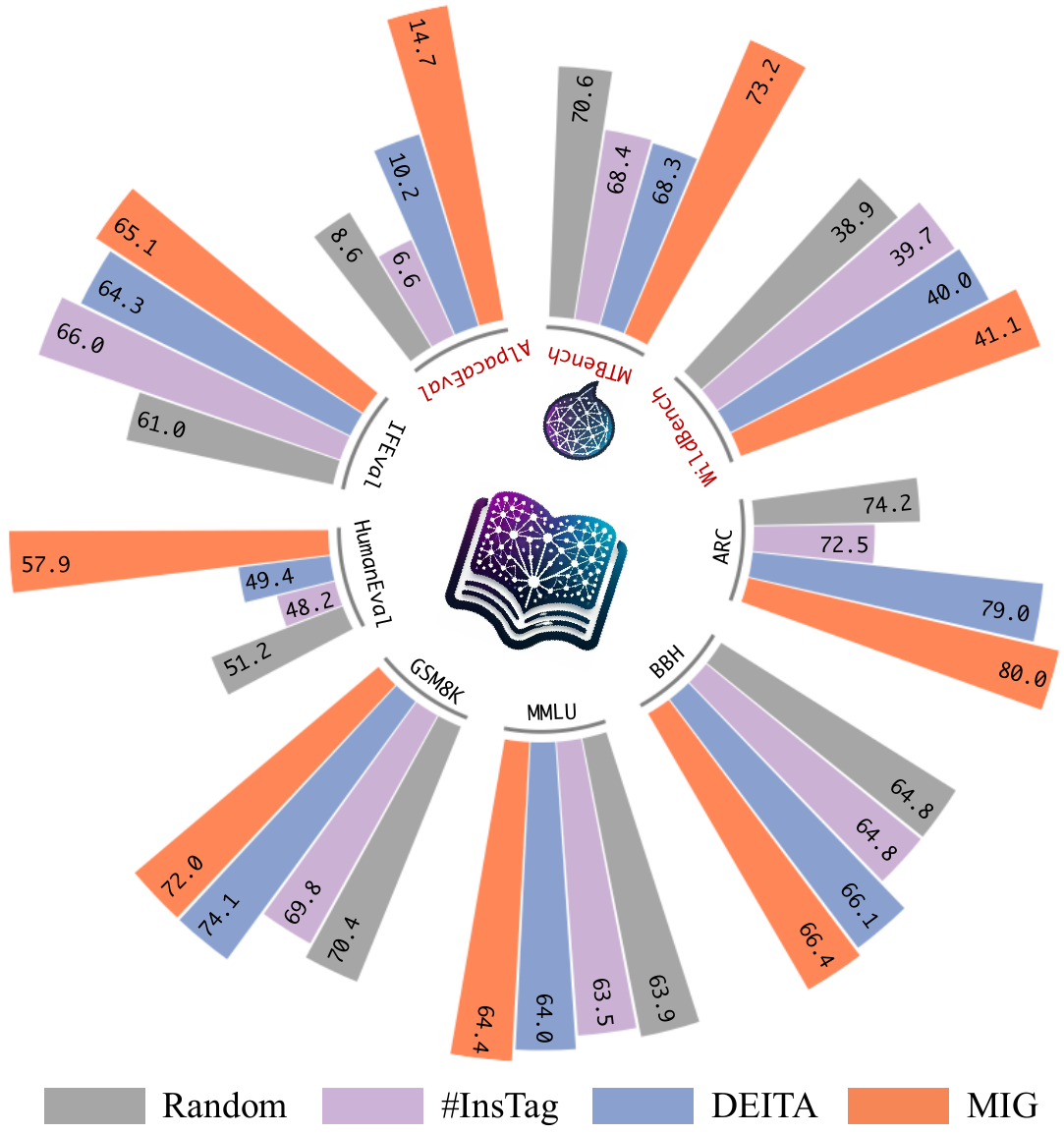}
    \caption{\small Comparison with different data selection methods~\cite{lu2024instag,liu2024what} on the Tulu3~\cite{lambert2024tulu3} pool using Llama3.1-8B~\cite{touvron2023llama}, evaluated on (black) knowledge-based benchmarks and (red) human-preference benchmarks. See details in Sec.~\ref{sec:main-results}.}
    \label{fig:teaser}
    \vspace{-16pt}
\end{figure}
%--------------------------------- Teaser Image ---------------------------------%

Large Language Models (LLMs) have shown remarkable capabilities in following human instructions in a wide range of tasks~\cite{wang2023openchat}.
Typically, LLMs first acquire general knowledge through large-scale pretraining and are subsequently refined through instruction tuning to better align with diverse human intentions~\cite{NEURIPS2020_1457c0d6,alpaca,touvron2023llama}.
Instruction tuning utilizes instruction-response pairs to guide base models toward more accurate and contextually appropriate responses.
Recent studies~\cite{NEURIPS2023_ac662d74,chen2024alpagasus} emphasize the critical role of data engineering in instruction tuning, highlighting data quality rather than quantity as the key to effective instruction tuning.
Notably, LIMA~\cite{NEURIPS2023_ac662d74} demonstrates that just 1000 high-quality, human-curated instructions can achieve performance comparable to substantially larger datasets.
However, manual curation of such datasets is inherently time-consuming and labor-intensive~\cite{vicuna2023}.

More recently, a line of work~\cite{chen2024alpagasus,lu2024instag,liu2024what,bukharin2023data} proposes automatic selections of optimal subsets from extensive data pools by defining desirable data characteristics.
These approaches~\cite{bukharin2023data,yu2024diversifyconquerdiversitycentricdata} posit that \textbf{quality} and \textbf{diversity} are crucial for an effective instruction-tuning dataset.
Data quality is defined from multiple perspectives, such as instruction complexity~\cite{lu2024instag,zhao-etal-2024-tree}, model perplexity and uncertainty~\cite{li-etal-2024-quantity}, or scores assigned by advanced external models~\cite{chen2024alpagasus,liu2024what}.
However, diversity remains less explicitly quantified, often addressed via heuristic methods such as maximizing label set coverage~\cite{lu2024instag}, reducing redundancy through diversity filters~\cite{liu2024what}, or enforcing fixed sample distributions per cluster~\cite{ge-etal-2024-clustering,yu2024diversifyconquerdiversitycentricdata}. 
This narrow focus on diversity only during later selection stages, without a comprehensive view of the entire dataset, diminishes the global diversity and representativeness of the sampled data.
% , thus undermining model performance and generalizability.
% 
Alternative methods~\cite{bukharin2023data} employing embedding-based facility location functions~\cite{cornuejols1983uncapicitated} quantify diversity but require computationally intensive iterative pairwise distance calculations, making them impractical for large datasets.
Additionally, distance-based clustering in the embedding space may fail to capture the semantic intent of complex instructions accurately.
To solve these issues, several essential questions are raised: 
1) How can we effectively quantify diversity in semantic space while balancing quality and diversity in dataset evaluation?
2) How can we efficiently select data based on such evaluations?

To this end, we propose an information-based measure for instruction-tuning datasets and introduce an efficient data selection algorithm that aims to \textbf{M}aximize the \textbf{I}nformation \textbf{G}ain (MIG). 
We model the semantic space as a label graph, with nodes representing labels and edges capturing semantic relationships.
Information in the dataset is distributed across this graph, with the total information being the sum of each label's information.
% the aggregate of the information from each label.
% 
Each data point contributes to its associated labels in proportion to its quality.
Thus, the information of each data point measures local data quality, while the total of all label information measures the global diversity of the dataset.
To balance quality and diversity, we apply a monotonically increasing but marginally diminishing function to compute label information, thereby promoting diversity and preventing excessive data concentration on particular labels.
To better model information distribution in semantic space, we propagate information along label graph edges to address semantic correlations and annotation biases.
Leveraging the submodularity of our proposed information-based dataset measurement, we implement an efficient greedy algorithm that iteratively selects data points that maximize the information gain according to the current state of the label graph.
% To efficiently select a dataset that maximizes the total information, 
% we leverage the submodularity of our dataset measurement and implement a greedy algorithm that iteratively selects data points that maximize the information gain according to the current state of the label graph.

Through extensive experiments across data pools~\cite{liu2024selectit,OpenHermes2.5,lambert2024tulu3} of varying quality and sizes, and LLMs of different families~\cite{touvron2023llama,jiang2023mistral7b,qwen2.5},
% namely Llama~\cite{touvron2023llama}, Mistral~\cite{jiang2023mistral7b} and Qwen~\cite{qwen2.5},
MIG consistently achieves superior performance on both human-preference and knowledge-based evaluations. 
As shown in Fig~\ref{fig:teaser}, on the Tulu3~\cite{lambert2024tulu3} pool with Llama3.1-8B as the base model, MIG achieves average improvements of \textbf{+1.49\%} on six knowledge-based benchmarks~\cite{clark2018think,suzgun2022challenging,hendrycks2021measuring,chen2021codex,cobbe2021gsm8k,zhou2023instruction} and \textbf{+1.96\%} on three human-preference benchmarks~\cite{zheng2023judging,dong-etal-2024-abilities,lin2024wildbench} compared to previous state-of-the-art data selection methods~\cite{liu2024what,bukharin2023data}.
When combining both evaluations, MIG achieves average improvements of \textbf{+2.20\%} compared to the second-best method~\cite{bukharin2023data}.
Notably, the model fine-tuned with 5\% Tulu3 data sampled by MIG outperforms the official SFT model trained on the full dataset by \textbf{+1.73\%} (average on nine benchmarks), with a substantial boost of \textbf{+4.59\%} in human-preference evaluations.
MIG also outperforms existing methods on the Openhermes2.5~\cite{OpenHermes2.5} and $X_{sota}$~\cite{lu2024instag,liu2024what}, further demonstrating its generalizability across different settings.
Additionally, MIG significantly enhances sampling efficiency, reducing sampling time by over 100-fold on the Tulu3 data pool compared to embedding-based methods.

In summary, our contributions are as follows: 
\begin{enumerate}[label={\bf {{$\bullet$}}}, leftmargin=*, topsep=0.5ex, itemsep=-0.5ex, partopsep=0.75ex, parsep=0.75ex, partopsep=0pt, wide, labelindent=0pt]
    \item We propose an information-based measurement for instruction-tuning datasets in semantic space. It quantifies quality and diversity within the information distributed across the semantic label graph.
    \item We introduce MIG, an efficient data selection algorithm that maximizes the information gain on the label graph iteratively.
    \item Extensive experiments on various data pools, base models, and benchmarks demonstrate the effectiveness and generalizability of MIG.
    The correlation between parameters in MIG and the attributes of sampled data is well studied.
\end{enumerate}
\section{Related Work}
\label{sec:related-work}

%--------------------------------- Method Image ---------------------------------%
\begin{figure*}[t]
    \centering
    \includegraphics[width=0.85\linewidth]{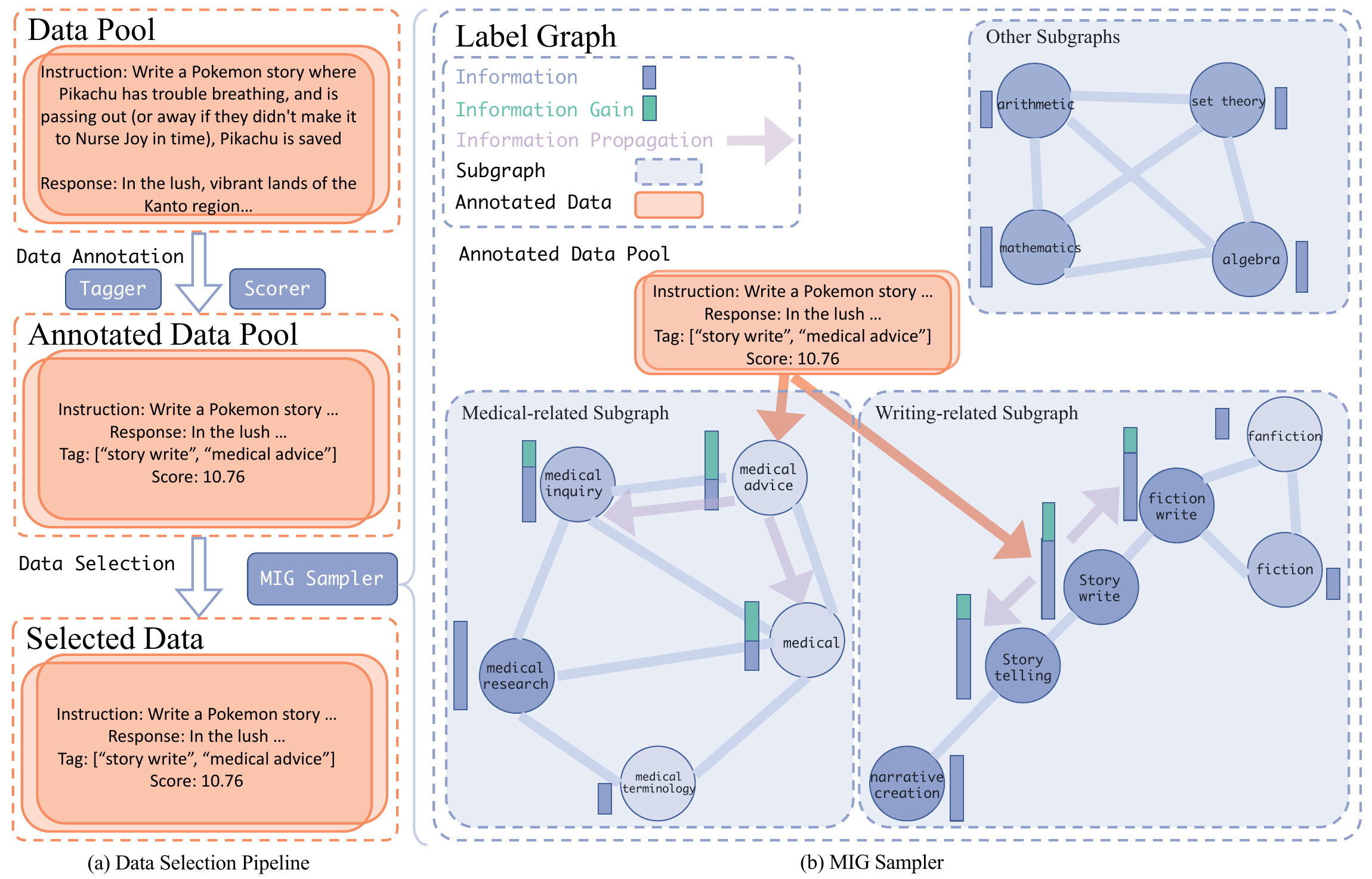}
    \caption{Illustration of (a) Data Selection Pipeline and (b) MIG Sampler. Given the raw data pool, our pipeline first applies a tagger and scorer to annotate data. Next, MIG constructs the label graph based on the label set and iteratively selects the data point that maximizes the information gain within the graph. The selected data are used for supervised fine-tuning (SFT) of LLMs.}
    \label{fig:arch}
    \vspace{-12pt}
\end{figure*}
%--------------------------------- Method Image ---------------------------------%

\noindent
\textbf{Data Selection for Instruction Tuning.}
% Instruction-tuning data can significantly enhance base LLMs.
% 
Recent studies~\cite{NEURIPS2023_ac662d74,chen2024alpagasus} indicate that increasing data quality and diversity rather than quantity effectively boosts instruction-following performance.
% has been shown to more effectively induce instruction-following abilities.
% 
Consequently, data selection methods aim to identify optimal subsets that meet such characteristics and generally fall into three categories:
% 
% These methodologies generally fall into two categories:
% 
(1) \textbf{Quality-based approaches} prioritize high-quality data points, where quality is defined through various perspectives, such as instruction complexity and response quality.
INSTRUCTMINING~\cite{cao2024instruction} identifies natural language metrics indicative of high-quality instruction data.
Instruction-Following Difficulty (IFD)~\cite{li-etal-2024-quantity} highlights inconsistencies between a model's anticipated responses and its self-generated outputs.
Nuggets~\cite{li2023one} measures quality based on the disparity between one-shot and zero-shot performance.
LESS~\cite{xia2024less} uses gradient features to select samples based on their similarity to a few representative examples.
SelectIT~\cite{liu2024selectit} selects high-quality data based on intrinsic uncertainty from token, sentence, and model levels.
Additionally, some methods employ external LLMs to assess data quality, such as 
ALPAGASUS~\cite{chen2024alpagasus}, which uses a well-designed prompt applied to ChatGPT to assess the quality of each data tuple.
(2) \textbf{Diversity-based approaches} aim to select data subsets with broad coverage of the data pool.
DiverseEvol~\cite{wu2023self} iteratively selects samples distant from previously selected data in the embedding space.
% maintains high diversity within selected subsets by progressively choosing data points that are distant from existing ones in the current embedding space of the model.
% 
ZIP~\cite{2024arXiv240706645Y} prioritizes subsets with low compression ratios, implicitly favoring diversity.
\textbf{(3) Comprehensive approaches}
strive to balance quality and diversity.
\textit{\#InsTag}~\cite{lu2024instag} employs ChatGPT to generate detailed open-ended tags for instructions and prioritize complex data with more tags while maximizing topic coverage.
DEITA~\cite{liu2024what} prioritizes high-quality data points while avoiding duplicates in the embedding space.
CaR~\cite{ge-etal-2024-clustering} and kMQ~\cite{yu2024diversifyconquerdiversitycentricdata} cluster data and sample high-quality points from each cluster.
However, these methods typically rely on heuristic rules rather than a unified quantitative metric to balance quality and diversity.
% lack a comprehensive and unified measure for subsets and rely on heuristic rules to balance quality and diversity. 

\noindent
\textbf{Submodular function for Diversity Measurement.}
Traditional submodular functions, such as facility location, graph cut, and log determinant, effectively quantify dataset diversity by identifying representative, non-redundant subsets.
% in subsets.
% % 
% Maximizing a submodular function helps identify representative, non-redundant subsets.
% , such as facility location, graph cut, or log determinant, 
% 
Leveraging this property, QDIT~\cite{bukharin2023data} measures diversity using the facility location function~\cite{cornuejols1983uncapicitated}, combining it linearly with quality scores.
Similarly, DPP~\cite{wang2024diversity} employs the log determinant distance to quantify subset diversity.
% and introduces a hyperparameter to balance diversity and quality. 
% 
Although this NP-hard problem can be approximated with a greedy algorithm following submodularity~\cite{nemhauser1978analysis,minoux2005accelerated}, embedding-based methods are inefficient at scale due to the high storage and computational costs of calculating high-dimensional pairwise distances.
To mitigate this issue, our custom dataset measurement is submodular, which justifies the use of a greedy strategy, while MIG samples data in a high-level semantic space, substantially reducing computational overhead.
\section{Method}
As shown in Fig.~\ref{fig:arch}(a), we begin by annotating the raw data pool with a tagger and scorer.
Next, MIG constructs a label graph to measure dataset information (Sec.~\ref{sec:info-meas}) and selects a subset for subsequent SFT by maximizing the information gain (Sec.~\ref{sec:mig-sampling}).

\subsection{Preliminary}
\noindent
\textbf{Task.}
Given a data pool $D_P$, a budget $N$, and an information measure $E(D)$ over any dataset $D$, the goal is to select a subset $D_S \subset D_P$ of size $N$ that maximizes $E(D)$. Formally,
\begin{equation}
\label{eq:data-selection}
\small
    D_S = \operatorname*{argmax}_{D \subset D_P,|D|=N} E(D)
\end{equation}

\noindent
\textbf{Data.}
Each data point is formed as:
\begin{equation}
\label{eq:d}
\small
    d_i = \{(q_i^j,r_i^j)_{j=1}^{M},L_i,s_i\}
\end{equation}
where $(q_i^j,r_i^j)_{j=1}^{M}$ represents $M$ rounds of query-response pairs used for training, $L_i$ is the set of labels (e.g., task category, knowledge domain, and other meta information) associated with $d_i$, and $s_i$ is the quality score.

\subsection{Information Measurement}
\label{sec:info-meas}
\noindent
\textbf{Label Graph.}
Previous studies~\cite{lu2024instag,ge-etal-2024-clustering,yu2024diversifyconquerdiversitycentricdata} assume that labels (including embedding-based clusters) are independent, ignoring the semantic relationships among them.
However, such label associations are crucial for accurately capturing the information distribution in semantic space.
% label-balance sampling.
% 
Intuitively, we can model labels as nodes, their associations as edges, and the intensity of associations as edge weights, thus modeling semantic space as an undirected weighted graph $G_L=(L, E_L)$, where $L$ represents the label set with a size of $K$ and $E_L$ represents edges. 
Specifically, we use label similarities as edge weights and remove edges whose weights are below a threshold $T$ to ensure computational efficiency.
Therefore, $E_L$ can be formed as a weighted adjacency matrix $W_L\in \mathbb{R}^{K\times K}$ with elements:
\begin{equation}
\small
    w_{pq}=\sigma[w(l_p,l_q) \geq T]\cdot w(l_p,l_q)
\end{equation}
where $w(l_p,l_q)$ represents textual similarity between label $l_p$ and $l_q$, and $\sigma(\cdot)$ yields $1$ when the input is evaluated as True.

\noindent
\textbf{Data Point Information.}
% We first define the information contributed by a single data point and then generalize it to the entire dataset.
% 
Under the label set $L$, a data point $d_i$ can be formed as a binary label vector with its associated labels $L_i$:
\begin{equation}
\small
    \mathbf{v}_i=\{v_k^i=\sigma(l_k\in L_i)\}_{k=1}^K
\end{equation}
The information of $d_i$ is distributed over $L_i$ and is proportional to its quality score $s_i$. Thus, the raw information of $d_i$ can be formed as: 
\begin{equation}
\small
    \mathbf{e}_i=s_i \cdot \mathbf{v}_i
\end{equation}
Semantic overlaps between labels and annotation-induced bias can lead to inaccurate information distribution. To address this, we introduce information propagation along the edges of the label graph, enabling a more accurate modeling of information distribution across the semantic space.
% Beyond directly contributing to the information of $L_i$, a data point also influences its neighboring labels through a propagation process along the edges of the label graph.
% 
Formally, the propagation from $l_p$ to $l_q$ is:
\begin{equation}
\label{eq:prop}
\small
    a_{pq} = \frac{\alpha w_{pq}}{w_{p}+\alpha \sum_{k, k\neq p} w_{pk}}
\end{equation}
where $w_p$ equals $1$ and $\alpha$ is a hyperparameter controlling the intensity of information propagation.
Let $A$ be the propagation matrix, then the propagated information vector of $d_i$ is:
\begin{equation}
\small
    \hat{\mathbf{e}}_i=A\mathbf{e}_i
\end{equation}

\noindent
\textbf{Dataset Information.}
% To promote a diverse distribution of labels within the label graph,
To balance quality and diversity within the label graph, we apply a monotonically increasing yet upper-convex function $\phi$ to compute the label information.
% instead of a simple summation.
% 
The marginally diminishing information gain is negatively correlated with the existing label information. Thus, information gains on labels with less information are prioritized.
Formally, the dataset information is:
% defined as:
\begin{equation}
\label{eq:dataset-measure}
\small
    E(D)=\Phi(\sum_{i\in D}A\mathbf{e}_i) = \Phi(A\sum_{i\in D}s_i\mathbf{v}_i)
\end{equation}
where $\Phi$ is a nonlinear transformation that applies $\phi$ element-wise to the input and then aggregates the results by summation.

\subsection{MIG Sampling}
\label{sec:mig-sampling}
% \textbf{MIG Sampling.}
Directly selecting $D_S$ from $D_P$ is computationally infeasible as the combination $C_{|D_P|}^{N}$ grows quickly.
Thus, as shown in Fig~\ref{fig:arch}(b), we follow the submodularity of $E(D)$ (detailed in Appx.~\ref{appx:proof}) and propose a greedy strategy, iteratively selecting the data point that yields the maximum information gain:
\begin{equation}
\label{eq:gain}
\small
    d_k=\operatorname*{argmax}_{d\in D_P^k}\{E(D_S^k\cup\{d\})-E(D_S^k)\}
\end{equation}
where $D_S^k$ and $D_P^k$ denote the selected subset and remaining candidate pool at iteration $k$.
We approximate the information gain in Eq.~\ref{eq:gain} via a gradient-based approach:
\begin{equation}
\small
    G_k = \frac{\partial E(D_S^k)}{\partial E} = A\Phi'(A\sum_{i\in D_S^k}\mathbf{e}_i)
\end{equation}
where $\Phi'$ represents the derivative of $\Phi$.
Thus, the selection process can be formed as:
\begin{equation}
\small
    d_k=\operatorname*{argmax}_{d\in D_P^k}G_k\mathbf{e}_d
\end{equation}
The sampling is detailed in Alg.~\ref{alg:sample}. Refer to Appx.~\ref{appx:implementation-details} for more implementation details of MIG.

\begin{algorithm}[t]
\caption{MIG Sampling}\label{alg:sample}
\KwData{Initial Data Pool $D_P$, Label Sets $L$, Sample Budget $N$}
\KwResult{The Sampled Dataset $D_S$}
Initialize Empty $D_S$\;
Initialize Propagation Matrix $A$\;
\While{$|D_S| < N$}{
    $G \gets A\Phi'(A\sum_{k\in D_S}E_k)$\;
    $d_i \gets \operatorname*{argmax}_{d\in D_P} GE_d$\;
    $D_S \gets D_S \cup \{d_i\}$\;
    $D_P \gets D_P \setminus \{d_i\}$\;
}
\Return{$D_S$}
\end{algorithm}
% \vspace{-16pt}
\begin{table*}[!t]
    \footnotesize
    \centering
    \caption{\footnotesize Comparison with data selection methods on the Tulu3 pool. HE denotes HumanEval, AE denotes AlpacaEvalv2, MT denotes MTBench, and Wild denotes WildBench. $\text{Avg}_{\text{obj}}$ and $\text{Avg}_{\text{sub}}$ represent the average of the normalized knowledge-based and human-preference benchmark scores, respectively. $\text{Avg}$ is the mean of $\text{Avg}_{\text{obj}}$ and $\text{Avg}_{\text{sub}}$. MIG achieves the best performance on $\text{Avg}_{\text{obj}}$, $\text{Avg}_{\text{sub}}$, and $\text{Avg}$ on all base models.}
    \label{tab:main-results}
    \resizebox{0.99\linewidth}{!}{
        \begin{tabular}{c|cc|ccccccc|cccc|c}
        \toprule[0.1em]
        Base Model & Method & Data Size & ARC & BBH & GSM & HE & MMLU & IFEval & $\text{Avg}_{\text{obj}}$ & AE & MT & Wild & $\text{Avg}_{\text{sub}}$ & $\text{Avg}$ \\
        \midrule
        \multirow{9}{*}{Llama3.1-8B} & Pool     &939K & 69.15 & 63.88 & 83.40 & 63.41 & 65.77 & 67.10 & 68.79 & 8.94 & 6.86 & -24.66 & 38.40 & 53.59 \\
        \\[-0.8em]
        \cline{2-15}
        \\[-0.8em]
        & Random   & 50K & 74.24 & 64.80 & 70.36 & 51.22 & 63.86 & 61.00 & 64.25 & 8.57 & \underline{7.06} & -22.15 & 39.36 & 51.81 \\
        & ZIP      & 50K & 77.63 & 63.00 & 52.54 & 35.98 & 65.00 & 61.00 & 59.19 & 6.71 & 6.64 & -32.10 & 35.69 & 47.44\\
        & IFD      & 50K & 75.93 & 63.56 & 61.03 & 49.39 & 64.39 & 53.60 & 61.32 & 12.30 & 7.03 & -20.20 & 40.83 & 51.08 \\
& \textit{\#InsTag}& 50K & 72.54 & 64.80 & 69.83 & 48.17 & 63.50 & \textbf{65.99} & 64.14 & 6.58 & 6.84 & -20.70 & 38.21 & 51.17 \\
        & DEITA    & 50K & 78.98 & 66.11 & \textbf{74.07} & 49.39 & 64.00 & 64.33 & \underline{66.15} & 10.19 & 6.83 & \underline{-19.95} & 39.50 & 52.83 \\
        & CaR      & 50K & 78.98 & \textbf{69.04} & 71.42 & 52.44 & \textbf{65.15} & 56.75 & 65.63 & 12.55 & 6.95 & -20.67 & 40.57 & 53.10 \\
        & QDIT     & 50K & \underline{79.66} & 65.42 & 70.74 & \underline{53.05} & \underline{65.06} & 57.30 & 65.21 & \textbf{15.78} & 6.76 & -20.56 & \underline{41.03} & \underline{53.12} \\ 
        \\[-0.8em]
        \cline{2-15}
        \\[-0.8em]
        & MIG      & 50K & \textbf{80.00} & \underline{66.39} & \underline{72.02} & \textbf{57.93} & 64.44 & \underline{65.06} & \textbf{67.64} & \underline{14.66} & \textbf{7.32} & \textbf{-17.77} & \textbf{42.99} & \textbf{55.32} \\
        \midrule
        \midrule
        \multirow{7}{*}{Mistral-7B-v0.3} & Random & 50K & 67.80 & 56.90 & \underline{66.34} & 42.07 & 60.34 & \textbf{65.43} & 59.81 & 5.84 & 6.84 & 
                                    -25.20 & 37.21 & 48.51 \\
                                    & ZIP    & 50K & 72.88 & 56.73 & 33.21 &  3.05 & 61.68 & 63.03 & 48.43 & 5.34 & 6.57 & -36.17 & 34.32 & 41.37 \\
                        & \textit{\#InsTag}  & 50K & \textbf{76.27} & 57.15 & \underline{66.34} & 40.85 & 61.80 & 63.22 & \underline{60.94} & 8.20 & 6.91 & -21.66 & 38.82 & \underline{49.88} \\
                                    & DEITA  & 50K & \underline{75.93} & 57.72 & 64.82 & 11.59 & 61.41 & 64.51 & 56.00 & 8.82 & 6.96 & -20.51 & 39.39 & 47.69 \\
                                    & CaR    & 50K & 64.41 & \underline{58.65} & 63.76 & 9.15  & \underline{61.95} & 55.64 & 52.26 & 11.93 & \underline{7.03} & \underline{-17.82} & 41.11 & 46.58 \\
                                    & QDIT   & 50K & 54.92 & \textbf{58.68} & 59.97 & \underline{42.68} & \textbf{62.46} & 58.23 & 56.16 & \textbf{15.03} & 6.84 & \textbf{-17.74} & \underline{41.52} & 48.84 \\
        \\[-0.8em]
        \cline{2-15}
        \\[-0.8em]
                                    & MIG    & 50K & 75.25 & 56.19 & \textbf{66.94} & \textbf{45.12} & 60.23 & \underline{64.70} & \textbf{61.41} & \underline{13.66} & \textbf{7.17} & -18.39 & \textbf{42.05} &  \textbf{51.73}\\
        \midrule
        \midrule
        \multirow{8}{*}{Qwen2.5-7B} & Pool   & 939K& 90.51 & 65.01 & 85.29 & 78.05 & 75.15 & 64.88 & 76.31 & 9.07 & 7.04 & -23.98 & 39.16 & 57.74 \\
        \\[-0.8em]
        \cline{2-15}
        \\[-0.8em]
        & Random & 50K & 85.42 & 63.87 & 80.74 & 79.27 & 73.81 & 58.04 & \underline{75.53} & 10.56 & 7.18 & \underline{-18.08} & 41.11 & 57.32 \\
        & ZIP    & 50K & 85.76 & 63.43 & 83.24 & 72.56 & 73.60 & 58.23 & 72.80 & 7.45  & 7.33 & -27.83 & 38.94 & 55.87 \\
&\textit{\#InsTag}& 50K& 88.81 & 63.03 & 84.61 & \textbf{81.10} & 73.50 & \underline{61.00} & 75.34 & 9.07 & \underline{7.52} & -18.80 & 41.62 & 58.48 \\
        & DEITA  & 50K & 89.15 & 63.22 & 86.13 & 79.27 & \underline{74.27} & 58.78 & 75.14 & 10.31 & 7.28 & -19.71 & 41.09 & 58.11 \\
        & CaR    & 50K & \textbf{91.86} & 65.60 & \textbf{87.64} & 77.44 & 73.97 & 50.28 & 74.47 & \underline{13.66} & 7.39 & -20.77 & \underline{42.39} & 58.43 \\
        & QDIT   & 50K & 89.83 & \textbf{69.34} & \underline{87.04} & \textbf{81.10} & \textbf{74.72} & 50.83 & 75.48 & \textbf{13.79} & 7.10 & -20.46 & 41.52 & \underline{58.50} \\
        \\[-0.8em]
        \cline{2-15}
        \\[-0.8em]
        & MIG    & 50K & \underline{90.51} & \underline{67.39} & 84.46 & 79.88 & 73.85 & \textbf{61.74} & \textbf{76.30} & 11.80 & \textbf{7.54} & \textbf{-14.49} & \textbf{43.32} & \textbf{59.81} \\ 
        \bottomrule[0.1em]
        \end{tabular}
    }
    \vspace{-8pt}
\end{table*}

\begin{table}[t]
\centering
\caption{\footnotesize Results on different data pools, Openhermes2.5 and $X_{sota}$, based on Llama3.1-8B. MIG outperforms all baselines across both data pools. 
Please refer to Table~\ref{tab:openhermes}~\ref{tab:deita} in Appx.~\ref{appx:detailed-results} for detailed scores on all benchmarks.
}
\label{tab:pools}
\resizebox{0.96\linewidth}{!}{
\begin{tabular}{c|cccc|cccc}
\toprule[0.1em]
\multirow{2}{*}{} & \multicolumn{4}{c|}{Openhermes2.5} & \multicolumn{4}{c}{$X_{sota}$} \\
\midrule
                  & Data Size & $\text{Avg}_{\text{sub}}$ & $\text{Avg}_{\text{obj}}$ & $\text{Avg}$ & Data Size & $\text{Avg}_{\text{sub}}$ & $\text{Avg}_{\text{obj}}$ & $\text{Avg}$ \\ 
\midrule
Pool               & 1M        & 36.91     & 61.49     & 49.20 & 306K      &  31.51    & 52.88 & 42.19 \\
\midrule
Random            & 50K       & 32.99     & 55.69     & 44.34 & 6K        &  29.94    & 49.69 & 39.81 \\
% ZIP               & 50K       &           &           & 6K        &  26.24    & 42.89 \\
% IFD               & 50K       &           &           & 6K        &  31.82    & 47.64 \\
\textit{\#InsTag} & 50K       & 36.23     & 54.12     & 45.17 & 6K        &  31.89    & 46.19 & 39.04 \\
DEITA             & 50K       & 36.80     & 57.36     & 47.08 & 6K        &  31.60    & 48.70 & 40.15 \\
CaR               & 50K       & 37.51     & 55.57     & 46.54 & 6K        &  31.86    & 48.43 & 40.15 \\
QDIT              & 50K       & 37.90     & 57.71     & 47.80 & 6K        &  32.52    & 49.10 & 40.81 \\
\midrule
MIG               & 50K       & \textbf{38.12}     & \textbf{58.30}    & \textbf{48.21} & 6K        &  \textbf{32.98}    & \textbf{50.63} & \textbf{41.80}\\
\bottomrule[0.1em]
\end{tabular}
}
\vspace{-12pt}
\end{table}

\section{Experiments}

\subsection{Setups}

\noindent
\textbf{Datasets.}
To investigate data selection across various scenarios and demonstrate the robustness of MIG, we use three distinct data pools: 
\begin{enumerate}[label={\bf {{$\bullet$}}}, leftmargin=*, topsep=0.5ex, itemsep=-0.5ex, partopsep=0.75ex, parsep=0.75ex, partopsep=0pt, wide, labelindent=0pt]
    \item Tulu3~\cite{lambert2024tulu3}: A large-scale, real-world SFT dataset presented by Ai2, containing million-level records across a wide variety of subjects, including mathematics, programming, and user dialogues.
    \item Openhermes2.5~\cite{OpenHermes2.5}: A dataset with over 1 million data points, sourced from 16 distinct origins, including MetaMath~\cite{yu2024metamath}, CamelAI~\cite{li2023camel}, and others.
    \item $X_{sota}$~\cite{lu2024instag,liu2024what}: A combined data pool consisting primarily of high-quality conversations from datasets such as WizardLM (Alpaca), WizardLM (ShareGPT), UltraChat~\cite{ding-etal-2023-enhancing}, and ShareGPT~\cite{vicuna2023}, totaling 300K data points.
\end{enumerate}

\noindent
\textbf{Benchmarks.}
We use both human-preference and knowledge-based benchmarks to evaluate model performance comprehensively. The evaluation is conducted using OpenCompass~\cite{2023opencompass}, with the average results reported as normalized scores on a percentage scale. Detailed evaluation settings are provided in Appx.~\ref{appx:evaluation-setup}.
\begin{enumerate}[label={\bf {{$\bullet$}}}, leftmargin=*, topsep=0.5ex, itemsep=-0.5ex, partopsep=0.75ex, parsep=0.75ex, partopsep=0pt, wide, labelindent=0pt]
\item Human-preference Benchmarks. We evaluate open-ended dialogue abilities using model-based evaluation metrics on three benchmarks:  AlpacaEvalv2~\cite{dubois2024length}, MTBench~\cite{zheng2023judging}, and WildBench~\cite{lin2024wildbench}.
\item Knowledge-based Benchmarks. We assess the factual knowledge, reasoning, coding, mathematical, and instruction-following abilities using automatic metrics on six benchmarks: ARC~\cite{clark2018think}, Big-Bench-Hard(BBH)~\cite{suzgun2022challenging}, MMLU~\cite{hendrycks2021measuring}, HumanEval~\cite{chen2021codex}, GSM8k~\cite{cobbe2021gsm8k}, and IFEval~\cite{zhou2023instruction}.
% ARC~\cite{clark2018think} and Big-Bench-Hard(BBH)~\cite{suzgun2022challenging} for natural language reasoning, MMLU~\cite{hendrycks2021measuring} for world knowledge, HumanEval~\cite{chen2021codex} for code generation, GSM8k~\cite{cobbe2021gsm8k} for mathematical reasoning, and IFEval~\cite{zhou2023instruction} for instruction-following.
\end{enumerate}

\noindent
\textbf{Baselines.}
We compare our methods against strong data selection approaches: random selection~\cite{xia2024rethinkingdataselectionscale}, IFD~\cite{li-etal-2024-quantity}, ZIP~\cite{2024arXiv240706645Y}, \textit{\#InsTag}~\cite{lu2024instag}, DEITA~\cite{liu2024what}, CaR~\cite{ge-etal-2024-clustering}, and QDIT~\cite{bukharin2023data}.
% Additionally,  is also considered a strong baseline, especially for knowledge-based evaluations.
% 
To replicate baselines on Tulu3 and Openhermes2.5, we adjust certain parameters to fit the large-scale datasets, as detailed in Appx~\ref{appx:baseline-settings}.

\noindent
\textbf{Training.}
We use LLaMA3.1-8B~\cite{touvron2023llama}, Mistral-7B-v0.3~\cite{jiang2023mistral7b}, and Qwen2.5-7B~\cite{qwen2.5} as our base models and fine-tune them using the Llama-Factory framework~\cite{zheng-etal-2024-llamafactory}. Please refer to Appx.~\ref{appx:training-recipes} for detailed training setup.

\subsection{Main Results}
\label{sec:main-results}
\noindent
\textbf{Main Comparison.} 
Table~\ref{tab:main-results} presents the performance of MIG and baselines across benchmarks.
% Table~\ref{tab:main-results} presents the performance of MIG for instruction data selection compared to several baselines across various benchmarks.
% 
All methods select 50K samples based on the grid search (Sec~\ref{sec:data-size}).
With Llama3.1-8B, MIG outperforms all baselines on most tasks, with average improvements of \textbf{+1.49\%} and \textbf{+1.96\%} over previous state-of-the-art selection methods on knowledge-based and human-preference evaluations, respectively.
MIG surpasses QDIT, the second-best method, by \textbf{+2.20\%} on overall $\text{Avg}$ score.
% When simultaneously considering both evaluation settings, MIG surpasses the second-best method, QDIT, by \textbf{+2.20\%} on $\text{Avg}$ score, highlighting the high quality and diversity of its sampled data.
% 
Notably, the model trained on 5\% data sampled by MIG outperforms the model trained on the full Tulu3 pool by \textbf{+4.59\%} on human-preference benchmarks while maintaining comparable knowledge-based performance.
% , a comprehensive and high-quality SFT training dataset directly applicable to real-world scenarios, on $\text{Avg}$.
% 
% Specifically, MIG delivers substantial gains in human-preference evaluations, with an average improvement of \textbf{+4.59\%} across three benchmarks while maintaining comparable performance on knowledge-based benchmarks.
% 
% Previous research~\cite{yuan2023rrhf,dong-etal-2024-abilities} indicates that mathematical capability improves with the increasing training dataset size without plateauing. This explains the underperformance of the model trained on 50K samples compared to the model trained on the full pool on the GSM8K benchmark.
% 
Additionally, MIG significantly outperforms embedding-based methods in sampling efficiency due to reduced computational overhead. Please refer to Table~\ref{tab:efficiency} in Appx.~\ref{appx:efficiency-analysis} for detailed sampling times and efficiency analysis.
% Additionally, among methods that balance quality and diversity, MIG demonstrates superior efficiency on large pools. It is more efficient than DEITA and QDIT, as it eliminates the need for iterative pairwise similarity calculations in the embedding space. For detailed sampling times and efficiency analysis, please refer to Table~\ref{tab:efficiency} in Appx.~\ref{appx:efficiency-analysis}.

\noindent
\textbf{Transferability on Models.}
Table~\ref{tab:main-results} presents results for Mistral-7B and Qwen2.5-7B.
MIG consistently surpasses baselines with $\text{Avg}$ improvements of \textbf{+1.85\%} and \textbf{+1.31\%}, respectively, demonstrating its robustness.
% To assess the generalizability of MIG, we additionally conduct experiments on Mistral-7B and Qwen2.5-7B. 
% 
% As shown in Table~\ref{tab:main-results}, MIG outperforms all baseline methods with $\text{Avg}$ improvements of \textbf{+1.85\%} and \textbf{+1.31\%} on these two base models.
% 
Notably, the second-best selection method varies among different base models, further demonstrating the generalizability of MIG. 
% Some strong baselines from Llama3.1-8B and Qwen2.5-7B experience performance degradation when applied to Mistral-7B.

\noindent
\textbf{Transferability on Data Pools.}
Table~\ref{tab:pools} presents results on different data pools with varying sizes and quality.
% We conduct experiments on Openhermes2.5~\cite{OpenHermes2.5} and $X_{sota}$~\cite{lu2024instag,liu2024what} to further evaluate the robustness of MIG.
% 
MIG consistently outperforms all baselines, achieving $\text{Avg}$ improvements of \textbf{+0.41\%} and \textbf{+0.99\%} over previous best methods, further demonstrating its generalizability.
% Results in Table~\ref{tab:pools} demonstrate that MIG consistently outperforms all baseline selection methods across different data pools and sample sizes, 
% 
% Specifically, MIG shows $\text{Avg}$ improvements of \textbf{+3.87\%} and \textbf{+1.99\%} compared to random selection on the two data pools, and \textbf{+0.41\%} and \textbf{+0.99\%} over previous SOTA methods.
% 
Notably, on $X_{sota}$, all baselines exhibit performance degradation on knowledge-based evaluations, consistent with the findings in ~\cite{xia2024rethinkingdataselectionscale}.
We hypothesize that quality metrics, such as DEITA scores and tag counts, are biased toward multi-round, long samples that enhance subjective chat abilities. 
However, samples in specific domains, such as math and code, are typically single-turn.
MIG mitigates this bias by effectively balancing quality and diversity.
% thus providing a more effective selection for domain-specific tasks.

\begin{figure}[t]
    \centering
    \includegraphics[width=0.9\linewidth]{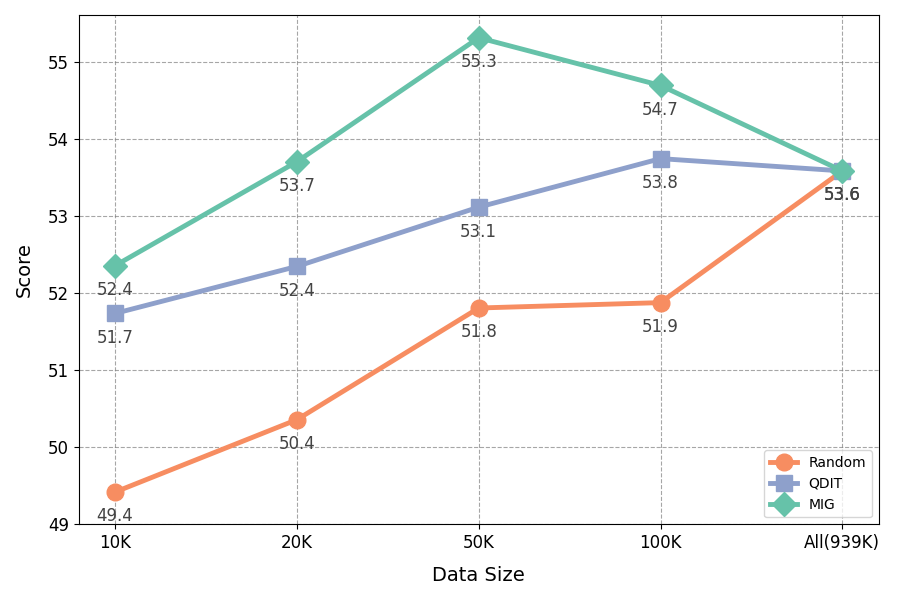}
    \captionof{figure}{\footnotesize Data scaling experiments on Tulu3 using Llama3.1-8B. The score reported here is the $\text{Avg}$ score.}
    \label{fig:data-scaling}
    \vspace{-12pt}
\end{figure}

\begin{figure*}[t]
    \centering
    \includegraphics[width=0.9\linewidth]{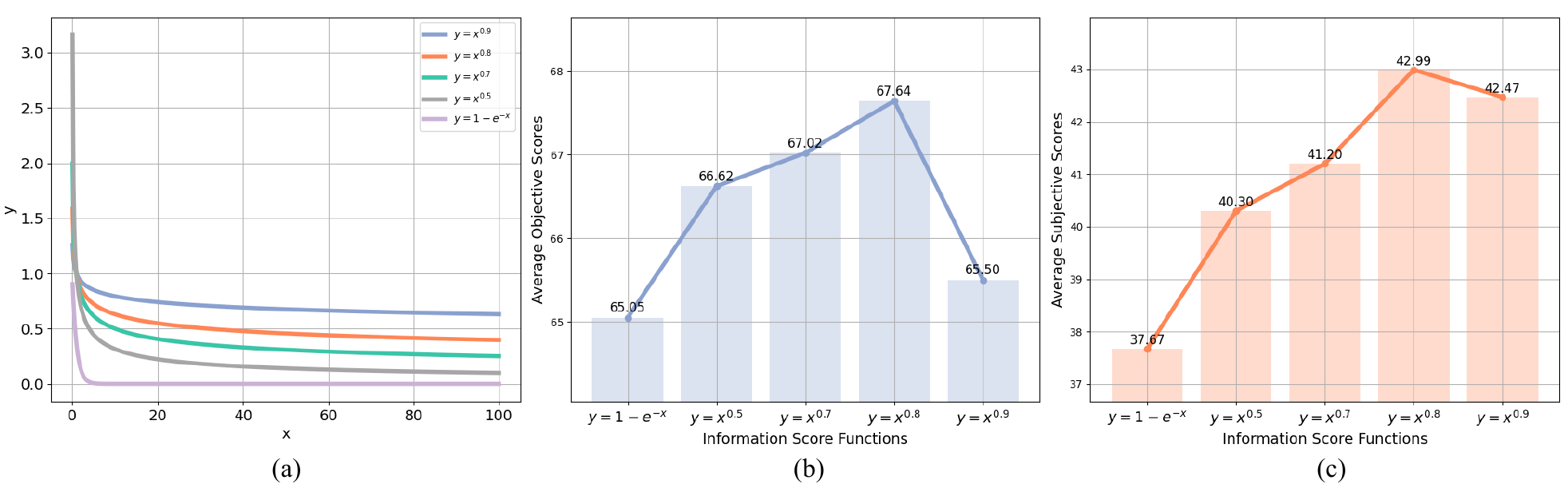}
    \captionof{figure}{\footnotesize (a) Derivative of Information Score Functions. (b) $\text{Avg}_{\text{obj}}$ on Different Information Score Functions. (c) $\text{Avg}_{\text{sub}}$ on Different Quality Scores.}
    \label{fig:info-score-funs}
    \vspace{-12pt}
\end{figure*}

\noindent
\textbf{Data Scaling.} 
We compare MIG with baseline methods across varying data budgets on the Tulu3 pool, using Llama3.1-8B as the base model.
As shown in Fig~\ref{fig:data-scaling}, MIG consistently delivers superior performance at each data budget, demonstrating its robust scalability.
Remarkably, MIG achieves comparable performance to the full dataset with only 20K samples, underscoring its efficiency. 
The observed initial increase and subsequent plateau in performance align closely with findings from previous works~\cite{li-etal-2024-quantity,liu2024what}, highlighting the importance of data selection.

\begin{figure}[t]
    \centering
    \includegraphics[width=0.85\linewidth]{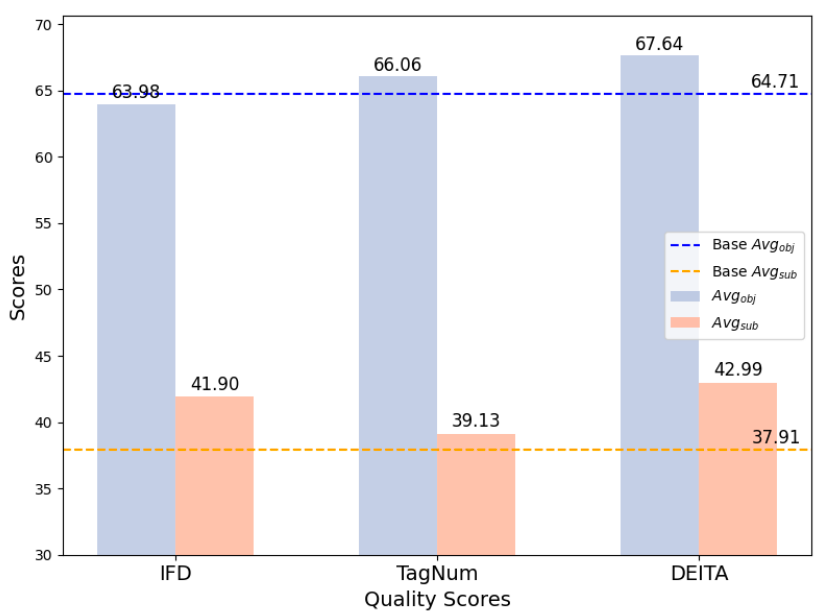}
    \captionof{figure}{\footnotesize Quantitative results on different quality metrics. DEITA scores achieve the best performance on both human-preference and knowledge-based evaluations.}
    \label{fig:quality}
    \vspace{-12pt}
\end{figure}

\begin{figure*}[t]
    \centering
    \includegraphics[width=0.9\linewidth]{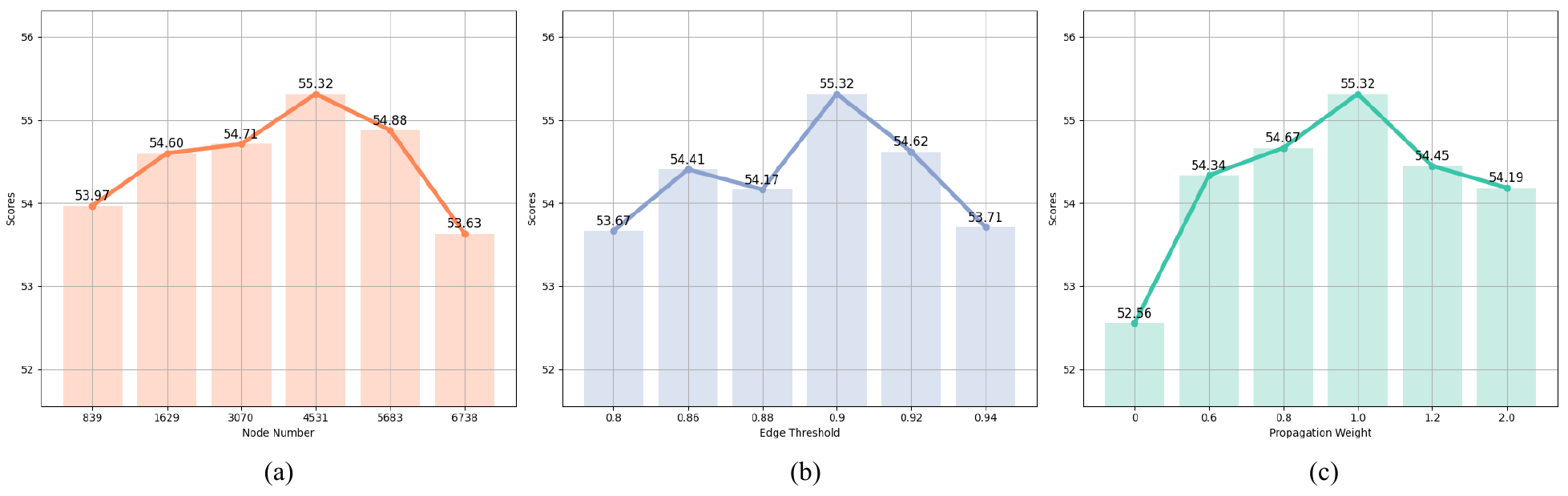}
    \captionof{figure}{\footnotesize Analysis of Parameters in the Label Graph. The reported score is the average of $\text{Avg}_{\text{sub}}$ and $\text{Avg}_{\text{obj}}$. 
    Please refer to Table~\ref{tab:node}~\ref{tab:edge}~\ref{tab:prop} in Appx.~\ref{appx:detailed-results} for detailed scores on all evaluated benchmarks. 
    (a) Comparison of various node counts (label set size) in the label graph. (b) Comparison of different edge thresholds, with a lower threshold indicating a dense graph. (c) Comparison of different propagation weights, where a smaller weight corresponds to weak propagation.}
    \label{fig:labelgraph-propagation}
    \vspace{-12pt}
\end{figure*}

\subsection{Analysis}
\label{sec:analysis}
\noindent
\textbf{Information Score Function $\Phi$.}
The information score function $\Phi$ is crucial in MIG sampling as it balances quality and diversity. 
Based on the principles outlined in Sec.~\ref{sec:info-meas}, $\Phi$ is expected to be monotonically increasing with a diminishing rate of increase.
In our experiments, we evaluate two candidate functions:
\begin{equation}
\small
    \Phi(x)=1-e^{-\alpha x} \quad (\alpha>0)
\end{equation}
\begin{equation}
\small
    \Phi(x)=x^{\alpha x} \quad (0<\alpha<1)
\end{equation}
Fig.~\ref{fig:info-score-funs}(a) compares the decreasing rate in the derivative of these functions under varying parameter settings.
Functions that decay rapidly tend to favor diverse label distributions as the information on any given label converges quickly. 
% 
% In contrast, slower decaying functions prioritize high-quality samples.
% 
Fig.~\ref{fig:info-score-funs}(b)(c) present the performance on different evaluations, with $\Phi(x)=x^{0.8}$ achieving the best results on human-preference and knowledge-based benchmarks, effectively balancing quality and diversity.

\noindent
\textbf{Quality Metrics.}
% Given the large scale of data pools in our experiments, reproducing methods from ~\cite{chen2024alpagasus,bukharin2023data}, which rely on computationally expensive external models such as ChatGPT for quality scoring, is impractical.
% 
% Therefore, we implement three alternative quality measurement approaches: the number of tags~\cite{lu2024instag}, the IFD score~\cite{li-etal-2024-quantity}, and the DEITA score~\cite{liu2024what}, to investigate their impact on information measurement.
We implement three alternative quality measurement approaches: the number of tags~\cite{lu2024instag}, the IFD score~\cite{li-etal-2024-quantity}, and the DEITA score~\cite{liu2024what}, to investigate their impact on information measurement.
Fig.~\ref{fig:quality} compares these three quality metrics with a baseline score that assigns a constant value to all samples.
The DEITA scores consistently outperform the other quality metrics in both evaluation settings.
Therefore, we adopt the DEITA scores as the default quality measurements for MIG.
% and other baseline selection methods.

\noindent
\textbf{Label Graph.}  
An essential question in MIG is how to determine an appropriate label graph, including its nodes (label set) and edges (label relationships).
Increasing the number of nodes leads to a more granular label set, thereby providing broader coverage of knowledge topics.
However, excessively large label sets inevitably include outliers or low-quality labels.
Similarly, increasing edge density between labels enhances the comprehensiveness of label relationships, but overly dense graphs may result in computational inefficiencies and noise from the embedding model.
There is no universally optimal solution, as the ideal label graph depends on the characteristics of the data pool and potentially other parameters in MIG.
To explore the relationship between the label graph and the downstream performance of trained models, we conduct an empirical experiment on the Tulu3 pool.
Fig.~\ref{fig:labelgraph-propagation}(a) shows the downstream performance from a set of node counts in the label graph, ranging from 839 to 6738, while Fig.~\ref{fig:labelgraph-propagation}(b) presents performance across varying edge densities, with thresholds between 0.8 and 0.94.
The observed trends align with our initial analysis, showing an unimodal performance curve in both experiments.
For the Tulu3 pool, the optimal label graph is achieved with a label set size of 4531 and an edge similarity threshold of 0.9.

\noindent
\textbf{Information Propagation.}
We conduct a series of experiments to study the impact of information propagation intensity in MIG sampling.
Appropriate information propagation results in accurate information distribution in the semantic space.
Specifically, we experiment with various values of $\alpha$ in Eq.~\ref{eq:prop}, where $\alpha$ is proportional to the intensity of information propagation.
% a higher $\alpha$ corresponds to stronger information propagation.
% and no propagation occurs when $\alpha=0$.
% 
Fig~\ref{fig:labelgraph-propagation}(c) shows that $\alpha=1.0$ yields the best performance, with $\text{Avg}$ improvement of \textbf{2.76} over the non-propagation.
It indicates that information propagation effectively improves the accuracy of information measurement on the label graph.
% Notably, results with information propagation significantly outperform those without, indicating that information propagation effectively captures the relationship between labels, thereby improving the accuracy of information measurement on the label graph.

\begin{table}[t]
\centering
\caption{\footnotesize Grid search of appropriate data size and training epochs on the Tulu3 pool. We report the AVG score here. 
% Please refer to the Table~\ref{tab:appx-grid} for detailed scores on all benchmarks.
}
\label{tab:grid}
\resizebox{0.96\linewidth}{!}{
\begin{tabular}{c|ccc|ccc}
\toprule[0.1em]
\multirow{2}{*}{} & \multicolumn{3}{c|}{Random} & \multicolumn{3}{c}{MIG} \\
\\[-0.8em]
\cline{2-7}
\\[-0.8em]
                  & Epoch2 & Epoch3 & Epoch4 & Epoch2 & Epoch3 & Epoch4 \\ 
\midrule
10K               & 46.76  & 49.42        & 50.39     & 49.13     &  52.36       & 51.11 \\
20K               & 48.23  & 50.36        & 51.08     & 51.18     &  53.71       & 53.14 \\
50K               & 49.78  &\textbf{51.81}& 50.68     & 52.88     &\textbf{55.32}& 55.14 \\
\bottomrule[0.1em]
\end{tabular}
}
\vspace{-12pt}
\end{table}

\noindent
\textbf{Grid Search.}
\label{sec:data-size}
To identify an appropriate data bucket and training epoch on the Tulu3 pool for the main comparison, we perform a grid search.
% using various data budgets and training epochs.
% 
% Table~\ref{tab:grid} shows that MIG consistently outperforms random selection across different data volumes, demonstrating its effectiveness.
% 
Results in Table~\ref{tab:grid} indicate 50K samples with three training epochs as optimal for Tulu3, consistently maximizing performance for MIG and random selection.
% For the default setting in the Tulu3 pool, we select 50K samples with three training epochs, as both random and MIG sampling achieve the best performance under this configuration.
\section{Conclusion}
In this paper, we propose a novel method for measuring instruction-tuning datasets in semantic space. 
We model the semantic space as a label graph and jointly evaluate data quality and diversity.
We introduce an upper-convex information score function to balance quality and diversity, and propose information propagation to capture the information distribution accurately
Building on the submodularity of such measurement, we propose MIG, an efficient sampling algorithm that iteratively selects samples to maximize the information gain on the label graph.
Extensive experiments across diverse data pools and base models validate the effectiveness and generalizability of MIG.
Our research bridges the gap between instance-level quality assessment and global dataset-level evaluation, offering a unified approach to dataset measurement.
We hope our results can inspire dataset measurement-guided data selection in the future.

\noindent
\textbf{Limitation.}
Currently, the parameters in MIG are static and depend on grid search to identify the optimal values, which can not be extensively explored. 
Future work could focus on developing methods to automatically determine the parameters in MIG, such as customizing the information score function for each label, to enhance the flexibility and scalability of MIG.

\bibliography{custom}

\begin{thebibliography}{49}
\providecommand{\natexlab}[1]{#1}

\bibitem[{Brown et~al.(2020)Brown, Mann, Ryder, Subbiah, Kaplan, Dhariwal, Neelakantan, Shyam, Sastry, Askell, Agarwal, Herbert-Voss, Krueger, Henighan, Child, Ramesh, Ziegler, Wu, Winter, Hesse, Chen, Sigler, Litwin, Gray, Chess, Clark, Berner, McCandlish, Radford, Sutskever, and Amodei}]{NEURIPS2020_1457c0d6}
Tom Brown, Benjamin Mann, Nick Ryder, Melanie Subbiah, Jared~D Kaplan, Prafulla Dhariwal, Arvind Neelakantan, Pranav Shyam, Girish Sastry, Amanda Askell, Sandhini Agarwal, Ariel Herbert-Voss, Gretchen Krueger, Tom Henighan, Rewon Child, Aditya Ramesh, Daniel Ziegler, Jeffrey Wu, Clemens Winter, Chris Hesse, Mark Chen, Eric Sigler, Mateusz Litwin, Scott Gray, Benjamin Chess, Jack Clark, Christopher Berner, Sam McCandlish, Alec Radford, Ilya Sutskever, and Dario Amodei. 2020.
\newblock Language models are few-shot learners.
\newblock In \emph{NIPS}.

\bibitem[{Bukharin et~al.(2024)Bukharin, Li, Wang, Yang, Yin, Li, Zhang, Zhao, and Jiang}]{bukharin2023data}
Alexander Bukharin, Shiyang Li, Zhengyang Wang, Jingfeng Yang, Bing Yin, Xian Li, Chao Zhang, Tuo Zhao, and Haoming Jiang. 2024.
\newblock Data diversity matters for robust instruction tuning.
\newblock In \emph{EMNLP}.

\bibitem[{Cao et~al.(2024{\natexlab{a}})Cao, Lam, Duan, Liu, Zhang, and Chen}]{cao2024compassjudger1allinonejudgemodel}
Maosong Cao, Alexander Lam, Haodong Duan, Hongwei Liu, Songyang Zhang, and Kai Chen. 2024{\natexlab{a}}.
\newblock Compassjudger-1: All-in-one judge model helps model evaluation and evolution.
\newblock \emph{arXiv preprint arXiv:2410.16256}.

\bibitem[{Cao et~al.(2024{\natexlab{b}})Cao, Kang, Wang, and Sun}]{cao2024instruction}
Yihan Cao, Yanbin Kang, Chi Wang, and Lichao Sun. 2024{\natexlab{b}}.
\newblock Instruction mining: Instruction data selection for tuning large language models.
\newblock In \emph{COLM}.

\bibitem[{Chen et~al.(2024)Chen, Li, Yan, Wang, Gunaratna, Yadav, Tang, Srinivasan, Zhou, Huang, and Jin}]{chen2024alpagasus}
Lichang Chen, Shiyang Li, Jun Yan, Hai Wang, Kalpa Gunaratna, Vikas Yadav, Zheng Tang, Vijay Srinivasan, Tianyi Zhou, Heng Huang, and Hongxia Jin. 2024.
\newblock Alpagasus: Training a better alpaca with fewer data.
\newblock In \emph{ICLR}.

\bibitem[{Chen et~al.(2021)Chen, Tworek, Jun, Yuan, de~Oliveira~Pinto, Kaplan, Edwards, Burda, Joseph, Brockman, Ray, Puri, Krueger, Petrov, Khlaaf, Sastry, Mishkin, Chan, Gray, Ryder, Pavlov, Power, Kaiser, Bavarian, Winter, Tillet, Such, Cummings, Plappert, Chantzis, Barnes, Herbert-Voss, Guss, Nichol, Paino, Tezak, Tang, Babuschkin, Balaji, Jain, Saunders, Hesse, Carr, Leike, Achiam, Misra, Morikawa, Radford, Knight, Brundage, Murati, Mayer, Welinder, McGrew, Amodei, McCandlish, Sutskever, and Zaremba}]{chen2021codex}
Mark Chen, Jerry Tworek, Heewoo Jun, Qiming Yuan, Henrique~Ponde de~Oliveira~Pinto, Jared Kaplan, Harri Edwards, Yuri Burda, Nicholas Joseph, Greg Brockman, Alex Ray, Raul Puri, Gretchen Krueger, Michael Petrov, Heidy Khlaaf, Girish Sastry, Pamela Mishkin, Brooke Chan, Scott Gray, Nick Ryder, Mikhail Pavlov, Alethea Power, Lukasz Kaiser, Mohammad Bavarian, Clemens Winter, Philippe Tillet, Felipe~Petroski Such, Dave Cummings, Matthias Plappert, Fotios Chantzis, Elizabeth Barnes, Ariel Herbert-Voss, William~Hebgen Guss, Alex Nichol, Alex Paino, Nikolas Tezak, Jie Tang, Igor Babuschkin, Suchir Balaji, Shantanu Jain, William Saunders, Christopher Hesse, Andrew~N. Carr, Jan Leike, Josh Achiam, Vedant Misra, Evan Morikawa, Alec Radford, Matthew Knight, Miles Brundage, Mira Murati, Katie Mayer, Peter Welinder, Bob McGrew, Dario Amodei, Sam McCandlish, Ilya Sutskever, and Wojciech Zaremba. 2021.
\newblock Evaluating large language models trained on code.
\newblock \emph{arXiv preprint arXiv:2107.03374}.

\bibitem[{Chiang et~al.(2023)Chiang, Li, Lin, Sheng, Wu, Zhang, Zheng, Zhuang, Zhuang, Gonzalez, Stoica, and Xing}]{vicuna2023}
Wei-Lin Chiang, Zhuohan Li, Zi~Lin, Ying Sheng, Zhanghao Wu, Hao Zhang, Lianmin Zheng, Siyuan Zhuang, Yonghao Zhuang, Joseph~E. Gonzalez, Ion Stoica, and Eric~P. Xing. 2023.
\newblock \href {https://lmsys.org/blog/2023-03-30-vicuna/} {Vicuna: An open-source chatbot impressing gpt-4 with 90\%* chatgpt quality}.

\bibitem[{Clark et~al.(2018)Clark, Cowhey, Etzioni, Khot, Sabharwal, Schoenick, and Tafjord}]{clark2018think}
Peter Clark, Isaac Cowhey, Oren Etzioni, Tushar Khot, Ashish Sabharwal, Carissa Schoenick, and Oyvind Tafjord. 2018.
\newblock Think you have solved question answering? try arc, the ai2 reasoning challenge.
\newblock \emph{arXiv preprint arXiv:1803.05457}.

\bibitem[{Cobbe et~al.(2021)Cobbe, Kosaraju, Bavarian, Chen, Jun, Kaiser, Plappert, Tworek, Hilton, Nakano, Hesse, and Schulman}]{cobbe2021gsm8k}
Karl Cobbe, Vineet Kosaraju, Mohammad Bavarian, Mark Chen, Heewoo Jun, Lukasz Kaiser, Matthias Plappert, Jerry Tworek, Jacob Hilton, Reiichiro Nakano, Christopher Hesse, and John Schulman. 2021.
\newblock Training verifiers to solve math word problems.
\newblock \emph{arXiv preprint arXiv:2110.14168}.

\bibitem[{Contributors(2023)}]{2023opencompass}
OpenCompass Contributors. 2023.
\newblock Opencompass: A universal evaluation platform for foundation models.
\newblock \url{https://github.com/open-compass/opencompass}.

\bibitem[{Cornu{\'e}jols et~al.(1983)Cornu{\'e}jols, Nemhauser, and Wolsey}]{cornuejols1983uncapicitated}
G{\'e}rard Cornu{\'e}jols, George Nemhauser, and Laurence Wolsey. 1983.
\newblock The uncapicitated facility location problem.
\newblock Technical report, Cornell University Operations Research and Industrial Engineering.

\bibitem[{Ding et~al.(2023)Ding, Chen, Xu, Qin, Hu, Liu, Sun, and Zhou}]{ding-etal-2023-enhancing}
Ning Ding, Yulin Chen, Bokai Xu, Yujia Qin, Shengding Hu, Zhiyuan Liu, Maosong Sun, and Bowen Zhou. 2023.
\newblock Enhancing chat language models by scaling high-quality instructional conversations.
\newblock In \emph{EMNLP}.

\bibitem[{Dong et~al.(2024)Dong, Yuan, Lu, Li, Xue, Liu, Wang, Yuan, Zhou, and Zhou}]{dong-etal-2024-abilities}
Guanting Dong, Hongyi Yuan, Keming Lu, Chengpeng Li, Mingfeng Xue, Dayiheng Liu, Wei Wang, Zheng Yuan, Chang Zhou, and Jingren Zhou. 2024.
\newblock How abilities in large language models are affected by supervised fine-tuning data composition.
\newblock In \emph{ACL}.

\bibitem[{Dubois et~al.(2024)Dubois, Galambosi, Liang, and Hashimoto}]{dubois2024length}
Yann Dubois, Bal{\'a}zs Galambosi, Percy Liang, and Tatsunori~B Hashimoto. 2024.
\newblock Length-controlled alpacaeval: A simple way to debias automatic evaluators.
\newblock \emph{arXiv preprint arXiv:2404.04475}.

\bibitem[{Ge et~al.(2024)Ge, Liu, Hu, Meng, Tao, Zhao, Xia, Li, Chen, Yang, Li, Xiao, and Zhu}]{ge-etal-2024-clustering}
Yuan Ge, Yilun Liu, Chi Hu, Weibin Meng, Shimin Tao, Xiaofeng Zhao, Mahong Xia, Zhang Li, Boxing Chen, Hao Yang, Bei Li, Tong Xiao, and JingBo Zhu. 2024.
\newblock Clustering and ranking: Diversity-preserved instruction selection through expert-aligned quality estimation.
\newblock In \emph{EMNLP}.

\bibitem[{Hahsler et~al.(2019)Hahsler, Piekenbrock, and Doran}]{hahsler2019dbscan}
Michael Hahsler, Matthew Piekenbrock, and Derek Doran. 2019.
\newblock dbscan: Fast density-based clustering with r.
\newblock \emph{Journal of Statistical Software}.

\bibitem[{Hendrycks et~al.(2021)Hendrycks, Burns, Basart, Zou, Mazeika, Song, and Steinhardt}]{hendrycks2021measuring}
Dan Hendrycks, Collin Burns, Steven Basart, Andy Zou, Mantas Mazeika, Dawn Song, and Jacob Steinhardt. 2021.
\newblock Measuring massive multitask language understanding.
\newblock In \emph{ICLR}.

\bibitem[{Jiang et~al.(2023)Jiang, Sablayrolles, Mensch, Bamford, Chaplot, de~las Casas, Bressand, Lengyel, Lample, Saulnier, Lavaud, Lachaux, Stock, Scao, Lavril, Wang, Lacroix, and Sayed}]{jiang2023mistral7b}
Albert~Q. Jiang, Alexandre Sablayrolles, Arthur Mensch, Chris Bamford, Devendra~Singh Chaplot, Diego de~las Casas, Florian Bressand, Gianna Lengyel, Guillaume Lample, Lucile Saulnier, Lélio~Renard Lavaud, Marie-Anne Lachaux, Pierre Stock, Teven~Le Scao, Thibaut Lavril, Thomas Wang, Timothée Lacroix, and William~El Sayed. 2023.
\newblock Mistral 7b.
\newblock \emph{arXiv preprint arXiv:2310.06825}.

\bibitem[{Lambert et~al.(2024)Lambert, Morrison, Pyatkin, Huang, Ivison, Brahman, Miranda, Liu, Dziri, Lyu, Gu, Malik, Graf, Hwang, Yang, Bras, Tafjord, Wilhelm, Soldaini, Smith, Wang, Dasigi, and Hajishirzi}]{lambert2024tulu3}
Nathan Lambert, Jacob Morrison, Valentina Pyatkin, Shengyi Huang, Hamish Ivison, Faeze Brahman, Lester James~V. Miranda, Alisa Liu, Nouha Dziri, Shane Lyu, Yuling Gu, Saumya Malik, Victoria Graf, Jena~D. Hwang, Jiangjiang Yang, Ronan~Le Bras, Oyvind Tafjord, Chris Wilhelm, Luca Soldaini, Noah~A. Smith, Yizhong Wang, Pradeep Dasigi, and Hannaneh Hajishirzi. 2024.
\newblock Tülu 3: Pushing frontiers in open language model post-training.
\newblock \emph{arXiv preprint arXiv:2411.15124}.

\bibitem[{Li et~al.(2023{\natexlab{a}})Li, Hammoud, Itani, Khizbullin, and Ghanem}]{li2023camel}
Guohao Li, Hasan Abed Al~Kader Hammoud, Hani Itani, Dmitrii Khizbullin, and Bernard Ghanem. 2023{\natexlab{a}}.
\newblock {CAMEL}: Communicative agents for ''mind'' exploration of large language model society.
\newblock In \emph{NIPS}.

\bibitem[{Li et~al.(2024{\natexlab{a}})Li, Zhang, He, Li, Zhao, Wang, Cheng, and Zhou}]{li-etal-2024-superfiltering}
Ming Li, Yong Zhang, Shwai He, Zhitao Li, Hongyu Zhao, Jianzong Wang, Ning Cheng, and Tianyi Zhou. 2024{\natexlab{a}}.
\newblock Superfiltering: Weak-to-strong data filtering for fast instruction-tuning.
\newblock In \emph{ACL}.

\bibitem[{Li et~al.(2024{\natexlab{b}})Li, Zhang, Li, Chen, Chen, Cheng, Wang, Zhou, and Xiao}]{li-etal-2024-quantity}
Ming Li, Yong Zhang, Zhitao Li, Jiuhai Chen, Lichang Chen, Ning Cheng, Jianzong Wang, Tianyi Zhou, and Jing Xiao. 2024{\natexlab{b}}.
\newblock From quantity to quality: Boosting {LLM} performance with self-guided data selection for instruction tuning.
\newblock In \emph{NAACL}.

\bibitem[{Li et~al.(2023{\natexlab{b}})Li, Hui, Xia, Yang, Yang, Zhang, Si, Liu, Liu, Huang et~al.}]{li2023one}
Yunshui Li, Binyuan Hui, Xiaobo Xia, Jiaxi Yang, Min Yang, Lei Zhang, Shuzheng Si, Junhao Liu, Tongliang Liu, Fei Huang, et~al. 2023{\natexlab{b}}.
\newblock One shot learning as instruction data prospector for large language models.
\newblock \emph{arXiv preprint arXiv:2312.10302}.

\bibitem[{Lin et~al.(2024)Lin, Deng, Chandu, Brahman, Ravichander, Pyatkin, Dziri, Bras, and Choi}]{lin2024wildbench}
Bill~Yuchen Lin, Yuntian Deng, Khyathi Chandu, Faeze Brahman, Abhilasha Ravichander, Valentina Pyatkin, Nouha Dziri, Ronan~Le Bras, and Yejin Choi. 2024.
\newblock Wildbench: Benchmarking llms with challenging tasks from real users in the wild.
\newblock \emph{arXiv preprint arXiv:2406.04770}.

\bibitem[{Liu et~al.(2024{\natexlab{a}})Liu, Liu, Wong, Li, Wang, Hu, and Zhang}]{liu2024selectit}
Liangxin Liu, Xuebo Liu, Derek~F. Wong, Dongfang Li, Ziyi Wang, Baotian Hu, and Min Zhang. 2024{\natexlab{a}}.
\newblock Select{IT}: Selective instruction tuning for {LLM}s via uncertainty-aware self-reflection.
\newblock In \emph{NIPS}.

\bibitem[{Liu et~al.(2024{\natexlab{b}})Liu, Zeng, He, Jiang, and He}]{liu2024what}
Wei Liu, Weihao Zeng, Keqing He, Yong Jiang, and Junxian He. 2024{\natexlab{b}}.
\newblock What makes good data for alignment? a comprehensive study of automatic data selection in instruction tuning.
\newblock In \emph{ICLR}.

\bibitem[{Lu et~al.(2024)Lu, Yuan, Yuan, Lin, Lin, Tan, Zhou, and Zhou}]{lu2024instag}
Keming Lu, Hongyi Yuan, Zheng Yuan, Runji Lin, Junyang Lin, Chuanqi Tan, Chang Zhou, and Jingren Zhou. 2024.
\newblock \#instag: Instruction tagging for analyzing supervised fine-tuning of large language models.
\newblock In \emph{ICLR}.

\bibitem[{Minoux(2005)}]{minoux2005accelerated}
Michel Minoux. 2005.
\newblock Accelerated greedy algorithms for maximizing submodular set functions.
\newblock In \emph{Optimization Techniques: Proceedings of the 8th IFIP Conference on Optimization Techniques W{\"u}rzburg, September 5--9, 1977}.

\bibitem[{Nemhauser et~al.(1978)Nemhauser, Wolsey, and Fisher}]{nemhauser1978analysis}
George~L Nemhauser, Laurence~A Wolsey, and Marshall~L Fisher. 1978.
\newblock An analysis of approximations for maximizing submodular set functions—i.
\newblock \emph{Mathematical programming}.

\bibitem[{Reimers and Gurevych(2019)}]{reimers-2019-sentence-bert}
Nils Reimers and Iryna Gurevych. 2019.
\newblock Sentence-bert: Sentence embeddings using siamese bert-networks.
\newblock In \emph{EMNLP}.

\bibitem[{Suzgun et~al.(2022)Suzgun, Scales, Sch{\"a}rli, Gehrmann, Tay, Chung, Chowdhery, Le, Chi, Zhou, , and Wei}]{suzgun2022challenging}
Mirac Suzgun, Nathan Scales, Nathanael Sch{\"a}rli, Sebastian Gehrmann, Yi~Tay, Hyung~Won Chung, Aakanksha Chowdhery, Quoc~V Le, Ed~H Chi, Denny Zhou, , and Jason Wei. 2022.
\newblock Challenging big-bench tasks and whether chain-of-thought can solve them.
\newblock \emph{arXiv preprint arXiv:2210.09261}.

\bibitem[{Taori et~al.(2023)Taori, Gulrajani, Zhang, Dubois, Li, Guestrin, Liang, and Hashimoto}]{alpaca}
Rohan Taori, Ishaan Gulrajani, Tianyi Zhang, Yann Dubois, Xuechen Li, Carlos Guestrin, Percy Liang, and Tatsunori~B. Hashimoto. 2023.
\newblock Stanford alpaca: An instruction-following llama model.
\newblock \url{https://github.com/tatsu-lab/stanford_alpaca}.

\bibitem[{Teknium(2023)}]{OpenHermes2.5}
Teknium. 2023.
\newblock \href {https://huggingface.co/datasets/teknium/OpenHermes-2.5} {Openhermes 2.5: An open dataset of synthetic data for generalist llm assistants}.

\bibitem[{Touvron et~al.(2023)Touvron, Lavril, Izacard, Martinet, Lachaux, Lacroix, Rozière, Goyal, Hambro, Azhar, Rodriguez, Joulin, Grave, and Lample}]{touvron2023llama}
Hugo Touvron, Thibaut Lavril, Gautier Izacard, Xavier Martinet, Marie-Anne Lachaux, Timothée Lacroix, Baptiste Rozière, Naman Goyal, Eric Hambro, Faisal Azhar, Aurelien Rodriguez, Armand Joulin, Edouard Grave, and Guillaume Lample. 2023.
\newblock Llama: Open and efficient foundation language models.
\newblock \emph{arXiv preprint arXiv:2302.13971}.

\bibitem[{Wang et~al.(2023{\natexlab{a}})Wang, Cheng, Zhan, Li, Song, and Liu}]{wang2023openchat}
Guan Wang, Sijie Cheng, Xianyuan Zhan, Xiangang Li, Sen Song, and Yang Liu. 2023{\natexlab{a}}.
\newblock Openchat: Advancing open-source language models with mixed-quality data.
\newblock \emph{arXiv preprint arXiv:2309.11235}.

\bibitem[{Wang et~al.(2023{\natexlab{b}})Wang, Yang, Huang, Yang, Majumder, and Wei}]{wang2023improving}
Liang Wang, Nan Yang, Xiaolong Huang, Linjun Yang, Rangan Majumder, and Furu Wei. 2023{\natexlab{b}}.
\newblock Improving text embeddings with large language models.
\newblock \emph{arXiv preprint arXiv:2401.00368}.

\bibitem[{Wang et~al.(2024)Wang, Shen, Guo, Stallone, Kim, Golland, and Panda}]{wang2024diversity}
Peiqi Wang, Yikang Shen, Zhen Guo, Matthew Stallone, Yoon Kim, Polina Golland, and Rameswar Panda. 2024.
\newblock Diversity measurement and subset selection for instruction tuning datasets.
\newblock \emph{arXiv preprint arXiv:2402.02318}.

\bibitem[{Wu et~al.(2023)Wu, Lu, Xu, Lin, Su, and Zhou}]{wu2023self}
Shengguang Wu, Keming Lu, Benfeng Xu, Junyang Lin, Qi~Su, and Chang Zhou. 2023.
\newblock Self-evolved diverse data sampling for efficient instruction tuning.
\newblock \emph{arXiv preprint arXiv:2311.08182}.

\bibitem[{Xia et~al.(2024{\natexlab{a}})Xia, Malladi, Gururangan, Arora, and Chen}]{xia2024less}
Mengzhou Xia, Sadhika Malladi, Suchin Gururangan, Sanjeev Arora, and Danqi Chen. 2024{\natexlab{a}}.
\newblock {LESS}: Selecting influential data for targeted instruction tuning.
\newblock In \emph{ICML}.

\bibitem[{Xia et~al.(2024{\natexlab{b}})Xia, Yu, Dang, Yang, Wu, Tian, Chang, and Lin}]{xia2024rethinkingdataselectionscale}
Tingyu Xia, Bowen Yu, Kai Dang, An~Yang, Yuan Wu, Yuan Tian, Yi~Chang, and Junyang Lin. 2024{\natexlab{b}}.
\newblock Rethinking data selection at scale: Random selection is almost all you need.
\newblock \emph{arXiv preprint arXiv:2410.09335}.

\bibitem[{Yang et~al.(2024)Yang, Yang, Zhang, Hui, Zheng, Yu, Li, Liu, Huang, Wei, Lin, Yang, Tu, Zhang, Yang, Yang, Zhou, Lin, Dang, Lu, Bao, Yang, Yu, Li, Xue, Zhang, Zhu, Men, Lin, Li, Xia, Ren, Ren, Fan, Su, Zhang, Wan, Liu, Cui, Zhang, and Qiu}]{qwen2.5}
An~Yang, Baosong Yang, Beichen Zhang, Binyuan Hui, Bo~Zheng, Bowen Yu, Chengyuan Li, Dayiheng Liu, Fei Huang, Haoran Wei, Huan Lin, Jian Yang, Jianhong Tu, Jianwei Zhang, Jianxin Yang, Jiaxi Yang, Jingren Zhou, Junyang Lin, Kai Dang, Keming Lu, Keqin Bao, Kexin Yang, Le~Yu, Mei Li, Mingfeng Xue, Pei Zhang, Qin Zhu, Rui Men, Runji Lin, Tianhao Li, Tingyu Xia, Xingzhang Ren, Xuancheng Ren, Yang Fan, Yang Su, Yichang Zhang, Yu~Wan, Yuqiong Liu, Zeyu Cui, Zhenru Zhang, and Zihan Qiu. 2024.
\newblock Qwen2.5 technical report.
\newblock \emph{arXiv preprint arXiv:2412.15115}.

\bibitem[{{Yin} et~al.(2024){Yin}, {Wu}, {Wang}, {Wang}, {Guo}, {Wang}, {Liu}, {Tang}, {Lian}, and {Chen}}]{2024arXiv240706645Y}
Mingjia {Yin}, Chuhan {Wu}, Yufei {Wang}, Hao {Wang}, Wei {Guo}, Yasheng {Wang}, Yong {Liu}, Ruiming {Tang}, Defu {Lian}, and Enhong {Chen}. 2024.
\newblock {Entropy Law: The Story Behind Data Compression and LLM Performance}.
\newblock \emph{arXiv preprint arXiv:2407.06645}.

\bibitem[{Yu et~al.(2024{\natexlab{a}})Yu, Jiang, Shi, YU, Liu, Zhang, Kwok, Li, Weller, and Liu}]{yu2024metamath}
Longhui Yu, Weisen Jiang, Han Shi, Jincheng YU, Zhengying Liu, Yu~Zhang, James Kwok, Zhenguo Li, Adrian Weller, and Weiyang Liu. 2024{\natexlab{a}}.
\newblock Metamath: Bootstrap your own mathematical questions for large language models.
\newblock In \emph{ICLR}.

\bibitem[{Yu et~al.(2024{\natexlab{b}})Yu, Chen, Ahmadian, and Fadaee}]{yu2024diversifyconquerdiversitycentricdata}
Simon Yu, Liangyu Chen, Sara Ahmadian, and Marzieh Fadaee. 2024{\natexlab{b}}.
\newblock Diversify and conquer: Diversity-centric data selection with iterative refinement.
\newblock \emph{arXiv preprint arXiv:2409.11378}.

\bibitem[{Zhao et~al.(2024)Zhao, Yu, Hui, Yu, Li, Huang, Zhang, and Li}]{zhao-etal-2024-tree}
Yingxiu Zhao, Bowen Yu, Binyuan Hui, Haiyang Yu, Minghao Li, Fei Huang, Nevin~L. Zhang, and Yongbin Li. 2024.
\newblock Tree-instruct: A preliminary study of the intrinsic relationship between complexity and alignment.
\newblock In \emph{COLING}.

\bibitem[{Zheng et~al.(2023)Zheng, Chiang, Sheng, Zhuang, Wu, Zhuang, Lin, Li, Li, Xing, Zhang, Gonzalez, and Stoica}]{zheng2023judging}
Lianmin Zheng, Wei-Lin Chiang, Ying Sheng, Siyuan Zhuang, Zhanghao Wu, Yonghao Zhuang, Zi~Lin, Zhuohan Li, Dacheng Li, Eric Xing, Hao Zhang, Joseph~E. Gonzalez, and Ion Stoica. 2023.
\newblock Judging {LLM}-as-a-judge with {MT}-bench and chatbot arena.
\newblock In \emph{NIPS}.

\bibitem[{Zheng et~al.(2024)Zheng, Zhang, Zhang, Ye, and Luo}]{zheng-etal-2024-llamafactory}
Yaowei Zheng, Richong Zhang, Junhao Zhang, Yanhan Ye, and Zheyan Luo. 2024.
\newblock {L}lama{F}actory: Unified efficient fine-tuning of 100+ language models.
\newblock In \emph{ACL}.

\bibitem[{Zhou et~al.(2023{\natexlab{a}})Zhou, Liu, Xu, Iyer, Sun, Mao, Ma, Efrat, Yu, YU, Zhang, Ghosh, Lewis, Zettlemoyer, and Levy}]{NEURIPS2023_ac662d74}
Chunting Zhou, Pengfei Liu, Puxin Xu, Srinivasan Iyer, Jiao Sun, Yuning Mao, Xuezhe Ma, Avia Efrat, Ping Yu, LILI YU, Susan Zhang, Gargi Ghosh, Mike Lewis, Luke Zettlemoyer, and Omer Levy. 2023{\natexlab{a}}.
\newblock Lima: Less is more for alignment.
\newblock In \emph{NIPS}.

\bibitem[{Zhou et~al.(2023{\natexlab{b}})Zhou, Lu, Mishra, Brahma, Basu, Luan, Zhou, and Hou}]{zhou2023instruction}
Jeffrey Zhou, Tianjian Lu, Swaroop Mishra, Siddhartha Brahma, Sujoy Basu, Yi~Luan, Denny Zhou, and Le~Hou. 2023{\natexlab{b}}.
\newblock Instruction-following evaluation for large language models.
\newblock \emph{arXiv preprint arXiv:2311.07911}.

\end{thebibliography}

\clearpage
\appendix

\section{Details of Experiments Setup}
\label{sec:exp-details}

\subsection{Baseline Settings}
\label{appx:baseline-settings}
The specific baseline settings in our experiments are as follows:
\begin{enumerate}[label={\bf {{$\bullet$}}}, leftmargin=*, topsep=0.5ex, itemsep=-0.5ex, partopsep=0.75ex, parsep=0.75ex, partopsep=0pt, wide, labelindent=0pt]
\item \textbf{Random}. A fixed random seed of 42 is used to ensure reproducibility.
\item \textbf{IFD}~\cite{li-etal-2024-quantity}. We follow the setting from ~\cite{li-etal-2024-superfiltering} to directly compute IFD scores using base LLMs for efficiency.
\item \textbf{ZIP}~\cite{2024arXiv240706645Y}. The default setting is applied across all data pools.
\item \textbf{\textit{\#InsTag}}~\cite{lu2024instag}. The open-released InsTagger is used to tag data from the pool, followed by tag normalization, which includes frequency filtering and semantic aggregation. The frequency threshold is set to $2$, as InsTagger is fully trained on valid normalized tags, resulting in 9471 tags. For semantic aggregation, we use E5-Mistral-7B-Instruct~\cite{wang2023improving} to generate tag embeddings, and the DBSCAN algorithm~\cite{hahsler2019dbscan} is applied with a semantic similarity threshold of 0.05, yielding 6738 tags.
\item DEITA~\cite{liu2024what}. For $X_{sota}$, we use the released sampled dataset. For the Openhermes2.5~\cite{OpenHermes2.5} and Tulu3~\cite{lambert2024tulu3} pools, we reproduce its method. Quality assessment is conducted using the released quality and complexity scorers. For \textit{Repr Filter}, we utilize Llama3.1-8B to obtain instance embeddings. A threshold of 0.9 is applied to the Tulu3 pool, and 0.95 is used for the Openhermes2.5 pool, as the 0.9 threshold results in insufficient samples for the latter.
\item CaR~\cite{ge-etal-2024-clustering}. We follow the default setting, using all-mpnet-base-v2~\cite{reimers-2019-sentence-bert} to obtain data embeddings, PCA to retain 95\% of dimensions, and k-Means clustering with the number of clusters as $k=\sqrt{n/2}$. For ranking, we use the released IQS model. We maintain the original ratio between $n1$ and $kn_2$ for different pools.
\item QDIT~\cite{bukharin2023data}. Embedding computation follows the default setting with all-mpnet-base-v2. Quality assessment uses the DEITA scores instead of ChatGPT. The hyperparameter $\alpha$ for balancing quality and diversity is set to 0.9 for $X_{sota}$ and 0.7 for the Tulu3 and Openhermes2.5 pools.
\end{enumerate}

\subsection{Implementation Details}
\label{appx:implementation-details}
The label set in MIG follows \textit{\#InsTag}~\cite{lu2024instag}, with variations in the label graph across different data pools. Specifically, 3059 tags are used for $X_{sota}$, 4531 for Tulu3, and 5166 for Openhermes2.5, with label set size positively correlated to pool size. 
E5-Mistral-7B-Instruct is used as the embedding model to compute label similarity, with a threshold set to 0.9.
Quality assessment in MIG uses the DEITA scores.
The information score function is set to $\phi(x)=x^{0.8}$, and the information propagation weight $\alpha$ is set to 1.

\subsection{Training Recipes}
\label{appx:training-recipes}
For experiments on $X_{sota}$, we follow the default settings from ~\cite{liu2024what}, using a batch size of 128, a learning rate of 2e-5, a warm ratio of 0.1, and a maximum input length of 2048.
For the Tulu3 pool, we adopt the settings from ~\cite{lambert2024tulu3}, with a batch size of 128, a learning rate of 5e-6, a warm ratio of 0.03, and a maximum input length of 4096.
For the Openhermes2.5 pool, we follow the settings from ~\cite{xia2024rethinkingdataselectionscale}, setting the batch size to 128, learning rates to 7e-6, a warm ratio of 0.01, and a maximum input length of 4096.

\subsection{Evaluation Setup}
\label{appx:evaluation-setup}
The evaluation of our experiments is implemented using OpenCompass~\cite{2023opencompass} with greedy inference to ensure uniform evaluation across all models. 

\noindent
\textbf{Human-preference Evaluations.} 
We use the open-source CompassJudger-1-32B~\cite{cao2024compassjudger1allinonejudgemodel} for human-preference evaluation.
As different benchmarks use different scoring metrics and ranges, we normalize the scores according to the following mapping:
\begin{enumerate}[label={\bf {{$\bullet$}}}, leftmargin=*, topsep=0.5ex, itemsep=-0.5ex, partopsep=0.75ex, parsep=0.75ex, partopsep=0pt, wide, labelindent=0pt]
\item AlpacaEvalv2~\cite{dubois2024length}: The score range is 0-100, requiring no special adjustment.
\item MTBench~\cite{zheng2023judging}: The score range is 0-10, which is mapped by multiplying by 10.
\item WildBench~\cite{lin2024wildbench}: The score range is -100-100, which is normalized by adding 100 and dividing by 2.
\end{enumerate}

\noindent
\textbf{Knowledge-based Evaluations.} 
We conduct evaluations using the following settings:
three-shot evaluation on BBH~\cite{suzgun2022challenging}, five-shot evaluation on MMLU~\cite{hendrycks2021measuring}, and
zero-shot evaluation on ARC~\cite{clark2018think} and GSM8K~\cite{cobbe2021gsm8k}.
For HumanEval~\cite{chen2021codex}, we report pass@1 results, and for IFEval~\cite{zhou2023instruction}, we provide strictly followed scores.

\begin{table}[t]
    \footnotesize
    \centering
    \caption{\footnotesize Efficiency comparison of different methods for 50K sampling from the Tulu3 pool, with timing measured on a single NVIDIA-L20Y.}
    \label{tab:efficiency}
    % \resizebox{0.5\linewidth}{!}{
        \begin{tabular}{c|cc}
        \toprule[0.1em]
        Method & GPU & Time \\
        \midrule
        Random & $\times$    & 0.09 \\
        ZIP~\cite{2024arXiv240706645Y}    & $\times$    & 53.99 \\
        IFD~\cite{li-etal-2024-quantity}    & $\times$    & 0.05 \\
        \#InsTag~\cite{lu2024instag} & $\times$    & 2.33 \\
        \midrule
        DEITA~\cite{liu2024what}  & $\checkmark$ & 81.56 \\
        QDIT~\cite{bukharin2023data}   & $\checkmark$ & 86.17 \\
        CaR~\cite{ge-etal-2024-clustering}    & $\checkmark$ & 0.85 \\
        MIG    & $\checkmark$ & 0.45 \\
        \bottomrule[0.1em]
        \end{tabular}
    % }
\end{table}

\section{Efficiency Analysis}
\label{appx:efficiency-analysis}
Table~\ref{tab:efficiency} presents the time used for 50K sampling on the Tulu3 pool.
Among methods that balance quality and diversity, MIG demonstrates the highest efficiency.
Notably, MIG outperforms QDIT~\cite{bukharin2023data} and DEITA~\cite{liu2024what} significantly, as it eliminates the need for iterative pairwise similarity computations in the embedding space.

\section{Theoretical Analysis of Greedy Strategy in MIG}
\label{appx:proof}

\subsection{Submodularity Analysis}

We first prove that our dataset measurement function $E(D)$, defined in Eq.~\ref{eq:dataset-measure}, is submodular. Specifically, for all $D\subseteq T$, and for all elements $e\notin T$, the following inequality holds:
\begin{equation}
\small
    E(D\cup e)-E(D)\geq E(T\cup e) - E(T)
\end{equation}

\begin{proof}
Let $\mathbf{z}(D) = A\sum_{i\in D}s_i\mathbf{v}_i$. The marginal gain from adding an element $e$ to $D$ is:
\begin{equation}
\small
    \Delta(D,e)=\sum_{k=1}^{n}[\phi(z_k(D)+\Delta_k(e))-\phi(z_k(D))]
\end{equation}
where $\Delta_k(e)=s_e(A\mathbf{v}_e)_k\geq0$ represents the incremental contribution of $e$ to the $k$-th component.

\noindent Since $\phi$ is monotonically increasing and concave, for any $\delta\geq0$ and $z'\geq z$, we have:
\begin{equation}
\small
\phi(z+\delta)-\phi(z)\geq \phi(z'+\delta)-\phi(z')
\end{equation}
Given $D\subseteq T$, we have:
\begin{equation}
\small
z_k(T)=z_k(D)+\sum_{i\in T\setminus D}s_i(A\mathbf{v}_i)_k\geq z_k(D)
\end{equation}
Thus, by the concavity property of $\phi$:
\begin{equation}
\begin{aligned}
\small
\phi(z_k(D)+\Delta_k(e))-\phi(z_k(D))\geq\\ \phi(z_k(T)+\Delta_k(e))-\phi(z_k(T))
\end{aligned}
\end{equation}
Summing over all components $k$, we get:
\begin{equation}
\small
\Delta(D,e)\geq \Delta(T,e)
\end{equation}
Thus, $E(D\cup e)-E(D)\geq E(T\cup e) - E(T)$, which satisfies the definition of submodularity.
\end{proof}

\subsection{Cardinality-Constrained Submodular Maximization}
Given that $E(D)$ is submodular, our data selection task, defined in Eq.~\ref{eq:data-selection}, constitutes a cardinality-constrained submodular maximization problem that is NP-complete. However, a greedy algorithm provides a well-established approximation guarantee, ensuring that $E(D^{greedy})\geq(1-\frac{1}{e})E(D^*)$, where $D^*$ represents the optimal solution~\cite{nemhauser1978analysis}. Assuming $P\neq NP$, this guarantee is the best achievable for polynomial-time algorithms.

\section{Detailed Results on Benchmarks}
\label{appx:detailed-results}
We provide detailed scores on full benchmarks in Table~\ref{tab:openhermes}~\ref{tab:deita}.
\begin{table*}[!t]
    \footnotesize
    \centering
    \caption{\footnotesize Full Results on the Openhermes2.5 Pool.}
    \label{tab:openhermes}
    \resizebox{0.96\linewidth}{!}{
        \begin{tabular}{c|ccccccc|cccc|c}
        \toprule[0.1em]
        Method & ARC & BBH & GSM8K & HumanEval & MMLU & IFEval & $\text{Avg}_{\text{obj}}$ & AlpacaEval & MTbench & Wildbench & $\text{Avg}_{\text{sub}}$ & AVG \\
        \midrule
        Pool      & 72.88 & 60.53 & 70.51 & 51.22 & 64.99 & 48.80 & 61.49 & 5.47 & 7.10 & -31.51 & 36.91 & 49.20 \\
        \midrule
        Random    & 75.25 & 60.20 & 51.40 & 50.00 & 51.23 & 46.03 & 55.69 & 4.72 & 6.63 & -44.12 & 32.99 & 44.34 \\
        InsTag    & 70.85 & 68.64 & 56.25 & 43.90 & 45.70 & 49.35 & 54.12 & 5.09 & 7.14 & -35.60 & 36.23 & 45.17 \\
        DEITA     & 69.83 & 61.85 & 60.96 & 46.95 & 58.01 & 46.58 & 57.36 & 7.83 & 6.94 & -33.69 & 36.80 & 47.08 \\
        CaR       & 62.71 & 63.73 & 55.42 & 44.51 & 64.37 & 42.70 & 55.57 & 7.33 & 7.09 & -31.43 & 37.51 & 46.54 \\
        QDIT      & 66.44 & 62.45 & 58.61 & 50.00 & 63.64 & 45.10 & 57.71 & 9.19 & 6.99 & -30.78 & 37.90 & 47.80 \\
        \midrule
        MIG       & 78.98 & 63.33 & 51.55 & 45.73 & 63.81 & 46.40 & 58.30 & 7.83 & 7.17 & -30.34 & 38.12 & 48.21 \\
        \bottomrule[0.1em]
        \end{tabular}
    }
\end{table*}

\begin{table*}[!t]
    \footnotesize
    \centering
    \caption{\footnotesize Full Results on the $X_{sota}$ Pool.}
    \label{tab:deita}
    \resizebox{0.96\linewidth}{!}{
        \begin{tabular}{c|ccccccc|cccc|c}
        \toprule[0.1em]
        Method & ARC & BBH & GSM8K & HumanEval & MMLU & IFEval & $\text{Avg}_{\text{obj}}$ & AlpacaEval & MTbench & Wildbench & $\text{Avg}_{\text{sub}}$ & AVG \\
        \midrule
        Pool      & 73.22 & 54.12 & 40.49 & 45.12 & 61.05 & 43.25 & 52.88 & 3.85 & 6.78 & -54.21 & 31.51 & 42.19 \\
        \midrule
        Random    & 61.02 & 58.12 & 32.07 & 42.69 & 62.31 & 41.96 & 49.69 & 3.60 & 6.34 & -54.39 & 29.94 & 39.81 \\
        InsTag    & 64.07 & 51.82 & 36.62 & 28.66 & 55.11 & 40.85 & 46.19 & 5.22 & 6.56 & -50.28 & 31.89 & 39.04 \\
        DEITA     & 71.86 & 50.82 & 27.67 & 40.24 & 63.36 & 38.26 & 48.70 & 4.22 & 6.48 & -48.44 & 31.60 & 40.15 \\
        CaR       & 72.88 & 48.90 & 20.92 & 46.95 & 62.68 & 38.26 & 48.43 & 5.22 & 6.51 & -49.46 & 31.86 & 40.15 \\
        QDIT      & 71.53 & 51.48 & 29.95 & 41.46 & 63.22 & 36.97 & 49.10 & 5.09 & 6.55 & -46.05 & 32.52 & 40.81 \\
        \midrule
        MIG       & 74.58 & 51.93 & 31.54 & 43.90 & 62.24 & 39.56 & 50.63 & 5.34 & 6.72 & -47.18 & 32.98 & 41.80 \\
        \bottomrule[0.1em]
        \end{tabular}
    }
\end{table*}

\begin{table*}[!t]
    \footnotesize
    \centering
    \caption{\footnotesize Full Results across Different Node Numbers.}
    \label{tab:node}
    \resizebox{0.96\linewidth}{!}{
        \begin{tabular}{c|ccccccc|cccc|c}
        \toprule[0.1em]
        Node Number & ARC & BBH & GSM8K & HumanEval & MMLU & IFEval & $\text{Avg}_{\text{obj}}$ & AlpacaEval & MTbench & Wildbench & $\text{Avg}_{\text{sub}}$ & AVG \\
        \midrule
        839      & 73.90 & 66.35 & 73.31 & 52.44 & 64.42 & 61.18 & 65.27 & 16.02 & 7.18 & -19.59 & 42.67 & 53.97 \\
        1629      & 76.27 & 66.22 & 72.18 & 56.71 & 64.71 & 58.96 & 65.84 & 15.16 & 7.40 & -18.15 & 43.36 &54.60 \\
        3070    & 78.31 & 66.80 & 72.25 & 55.49 & 65.03 & 63.22 & 66.85 & 14.04 & 7.25 & -17.63 & 42.58 & 54.71 \\
        4531     & 80.00 & 66.39 & 72.02 & 57.93 & 64.44 & 65.06 & 67.64 & 14.66 & 7.32 & -17.77 & 42.99 & 55.32 \\
        5683     & 83.73 & 66.47 & 72.55 & 55.49 & 64.22 & 64.14 & 67.77 & 12.55 & 7.35 & -20.15 & 41.99 & 54.88 \\
        6738     & 76.27 & 65.92 & 70.13 & 52.44 & 64.78 & 64.14 & 65.61 & 11.30 & 7.36 & -19.87 & 41.65 & 53.63 \\
        \bottomrule[0.1em]
        \end{tabular}
    }
\end{table*}

\begin{table*}[!t]
    \footnotesize
    \centering
    \caption{\footnotesize Full Results across Different Edge Densities.}
    \label{tab:edge}
    \resizebox{0.96\linewidth}{!}{
        \begin{tabular}{c|ccccccc|cccc|c}
        \toprule[0.1em]
        Node Number & ARC & BBH & GSM8K & HumanEval & MMLU & IFEval & $\text{Avg}_{\text{obj}}$ & AlpacaEval & MTbench & Wildbench & $\text{Avg}_{\text{sub}}$ & AVG \\
        \midrule
        0.8      & 77.97 & 65.69 & 70.96 & 54.88 & 64.59 & 63.96 & 66.34 & 11.3 & 7.24 & -21.45 & 40.99 & 53.67 \\
        0.86     & 80.68 & 65.72 & 71.49 & 55.49 & 64.15 & 62.85 & 66.73 & 14.04 & 7.21 & -19.77 & 42.08 & 54.41 \\
        0.88    & 76.95 & 67.77 & 72.02 & 57.93 & 64.99 & 62.85 & 67.09 & 13.42 & 7.04 & -20.17 & 41.25 & 54.17 \\
        0.9     & 80.00 & 66.39 & 72.02 & 57.93 & 64.44 & 65.06 & 67.64 & 14.66 & 7.32 & -17.77 & 42.99 & 55.32\\
        0.92     & 83.73 & 66.39 & 71.80 & 58.54 & 64.35 & 65.43 & 68.37 & 12.55 & 6.97 & -19.29 & 40.87 & 54.62 \\
        0.94     & 78.31 & 65.86 & 70.36 & 56.10 & 64.81 & 64.14 & 66.60 & 10.68 & 7.15 & -19.42 & 40.82 & 53.71 \\
        \bottomrule[0.1em]
        \end{tabular}
    }
\end{table*}

\begin{table*}[!t]
    \footnotesize
    \centering
    \caption{\footnotesize Full Results across Different Propagation Weights.}
    \label{tab:prop}
    \resizebox{0.96\linewidth}{!}{
        \begin{tabular}{c|ccccccc|cccc|c}
        \toprule[0.1em]
        Node Number & ARC & BBH & GSM8K & HumanEval & MMLU & IFEval & $\text{Avg}_{\text{obj}}$ & AlpacaEval & MTbench & Wildbench & $\text{Avg}_{\text{sub}}$ & AVG \\
        \midrule
        0      & 74.24 & 65.39 & 71.04 & 49.39 & 64.62 & 63.59 & 64.71 & 10.81 & 7.09 & -21.01 & 40.40 & 52.56 \\
        0.6       & 80.68 & 66.78 & 72.33 & 54.88 & 64.66 & 63.40 & 67.12 & 13.29 & 7.13 & -19.87 & 41.55 & 54.34 \\
        0.8    & 78.98 & 66.46 & 73.54 & 54.88 & 64.83 & 64.14 & 67.14 & 14.04 & 7.14 & -17.71 & 42.19 & 54.67 \\
        1.0     & 80.00 & 66.39 & 72.02 & 57.93 & 64.44 & 65.06 & 67.64 & 14.66 & 7.32 & -17.77 & 42.99 & 55.32 \\
        1.2     & 80.00 & 67.42 & 73.39 & 52.44 & 64.89 & 64.14 & 67.04 & 12.42 & 7.20 & -17.72 & 41.85 & 54.45 \\
        2.0     & 81.36 & 67.03 & 70.36 & 54.88 & 64.86 & 63.96 & 67.08 & 13.54 & 7.06 & -20.49 & 41.30 & 54.19 \\
        \bottomrule[0.1em]
        \end{tabular}
    }
\end{table*}

\end{document}